\documentclass[
twocolumn,
hf,
]{ceurart}

\usepackage[numbers]{natbib}
\usepackage{appendix}
\usepackage{lastpage}
\usepackage{chngcntr}  % For changing counter settings
\usepackage{float} 
\restylefloat{figure}  % Restyling the figure environment to apply float package settings
\usepackage[vlined,linesnumbered]{algorithm2e}
\SetKwInput{KwInput}{Input} % Set the Input
\usepackage{amsmath,amsfonts,amssymb,mathrsfs}
\usepackage{xcolor}
\usepackage{rotating}
\usepackage{booktabs} % For \toprule, \midrule, and \bottomrule
\usepackage{threeparttable}
\usepackage{subcaption}
\usepackage[normalem]{ulem}
\usepackage{mathrsfs}
\usepackage{newtxtext,newtxmath} % Modern Times Roman-like fonts

\usepackage{amssymb}

\def\tsc#1{\csdef{#1}{\textsc{\lowercase{#1}}\xspace}}
\tsc{WGM}
\tsc{QE}
\tsc{EP}
\tsc{PMS}
\tsc{BEC}
\tsc{DE}
\begin{document}

\title [mode = title]{A Review on Sound Source Localization in Robotics: Focusing on Deep Learning methods} 

% Authors

\author[1]{Reza Jalayer}[type=editor,
   auid=000,bioid=1,
   orcid=0000-0003-3440-5658]
\ead{reza.jalayer@polimi.it}
\cormark[1]

\author[2]{Masoud Jalayer}[type=editor,
   auid=000,bioid=1,
   orcid=0000-0001-8013-8613]

\author[3]{Amirali Baniasadi}[type=editor,
   auid=000,bioid=1,
   % orcid=0000-0001-8013-8613]
   ]

\address[1]{Department of Management, Economics and Industrial Engineering, Politecnico di Milano, Via Lambruschini 24/b, 20156, Milan, Italy}
% \address[2]{Department of Mechanical Engineering, University of California at Berkeley, Berkeley, CA 94709, USA}
\address[2]{Department of Materials and Mechanical Engineering, University of Turku, Vesilinnantie 5, Turku, 20014, Finland}
\address[3]{Department of Electrical and Computer Engineering, University of Victoria, Victoria, BC V8P 5C2, Canada}

% Define the conference
\conference{ArXiv}

\begin{abstract}
Sound-source localization (SSL) adds a spatial dimension to auditory perception, allowing a system to pinpoint the origin of speech, machinery noise, warning tones, or other acoustic events, capabilities that facilitate robot navigation, human–machine dialogue, and condition monitoring.
% Existing surveys offer valuable historical context but focus on generic audio applications and overlook the robotic constraints, and most predate the latest deep learning models. This review closes that gap with a robotics-centered synthesis and focusing more on deep learning advancements. We begin with well-established methods, e.g., Time Difference of Arrival (TDOA), beamforming, Generalized Cross-Correlation(GCC), and Multiple Signal Classification (MUSIC), and then focus on machine-learning and deep learning-based methods, considering works on simple, convolutional, and recurrent neural networks (NNs, CNNs, and RNNs) and recent attention-based architectures as well.
While existing surveys provide valuable historical context, they typically address general audio applications and do not fully account for robotic constraints or the latest advancements in deep learning. This review addresses these gaps by offering a robotics-focused synthesis, emphasizing recent progress in deep learning methodologies. We start by reviewing classical methods such as Time Difference of Arrival (TDOA), beamforming, Steered-Response Power (SRP), and subspace analysis. Subsequently, we delve into modern machine-learning (ML) and deep learning (DL) approaches, discussing traditional ML and neural networks (NNs), convolutional neural networks (CNNs), convolutional recurrent neural networks (CRNNs), and emerging attention-based architectures. The data and training strategy that are the two cornerstones of DL-based SSL are explored.
Studies are further categorized by robot types and application domains to facilitate researchers in identifying relevant work for their specific contexts. Finally, we highlight the current challenges in SSL works in general, regarding environmental robustness, sound source multiplicity, and specific implementation constraints in robotics, as well as data and learning strategies in DL-based SSL. Also, we sketch promising directions to offer an actionable roadmap toward robust, adaptable, efficient, and explainable DL-based SSL for next-generation robots.
\end{abstract}


\maketitle

\section{Introduction}

Sound source localization (SSL) is the task of estimating the location or direction of a sound-emitting source relative to the sensor (microphone array).
In robotic systems, SSL serves as a crucial component of robot audition, greatly enhancing a robot’s perceptual capabilities \cite{rascon2017localization}. 
An accurate SSL module enables a robot to orient its sensors towards the active speaker, disambiguate simultaneous talkers, or navigate autonomously towards an event that is audible but not visible.
Typical use cases range from domestic assistance, where a robot must detect who is issuing verbal commands \cite{jo2025sound,korayem2021design} as a part of broader task of speech recognition, to others applications in industrial manufacturing \cite{jalayer2023conceptual}, condition monitoring \cite{lv2024overview,lv2024motor}, autonomous vehicles \cite{marques2022microphone}, aerial robots \cite{yamada2020sound,yamamoto2024implementation}, search-and-rescue \cite{latif2015sound,zhang2019sound,mae2017sound}, and security robots \cite{park2008design,han2020research}, where detecting and localizing specific sounds, alarms, cries, or other auditory events is important for decision making.

Alternative localization in robotics can be done by vision, Light Detection And Ranging (LiDAR), Wi-Fi, Bluetooth, Global Positioning Systems (GPS), InfraRed (IR), Radio Frequency Identification (RFID). However, each of the aforementioned techniques has its pros and cons \cite{obeidat2021review}.
For example, Cameras are susceptible to occlusion, darkness, glare, or privacy constraints \cite{tarokh2010vision,hall2020robotic}; LiDARs and Wi-Fi based techniques are highly accurate in localization but expensive \cite{belkin2021real,yu2024wifi}; IR-based and Bluetooth localization are only accurate in specific conditions such that IR signals are easily obscured by physical obstacles and Bluetooth is range-limited; GPS is rendered unreliable by presence of multiple building walls \cite{wahab2022indoor} and RFID-based localization suffers from interference between readers in presence of multiple RFID tags and readers in the scene \cite{alfurati2018performance}.
Acoustic waves, in contrast, propagate around obstacles and in complete darkness, furnishing a complementary channel that often operates beyond the line of sight of on-board cameras. SSL is therefore not a rival but a synergistic partner to these modalities, enriching multi-sensor fusion and improving the robustness of perceptual pipelines.

Despite these advantages, sound-based localization is notoriously sensitive to microphone geometry, environmental reverberation and noise, and the presence of multiple simultaneous sound emitters. Early robotic SSL systems, dating back to the Squirt robot in 1989 \cite{flynn1989squirt}, relied on classical signal-processing algorithms \cite{rascon2017localization}.  These techniques model microphone spacing, speed of sound, and narrow-band free-field assumptions analytically; they achieve good performance in anechoic rooms or with a single talker but degrade quickly under strong reverberation, diffuse noise, and source motion.
The last decade has witnessed a paradigm shift toward data-driven learning, mirroring breakthroughs in computer vision and speech recognition.  Convolutional neural networks (CNNs), recurrent and convolutional–recurrent hybrids (CRNNs), residual and densely connected variants, and more recently attention-based transformers have demonstrated a remarkable ability to learn spatial features directly from raw waveforms or spectrograms \cite{grumiaux2022survey}.  By optimising end-to-end on millions of synthetic or recorded room-impulse responses, deep networks implicitly compensate for the challenging acoustic conditions such as multiple sound sources in a noisy and reverberant environment.  Moreover, embedded advanced edge computing now allows real-time inference on mobile platforms, making deep-learning SSL a viable solution rather than a laboratory curiosity.  For robotics in particular, the combination of low-cost microphone arrays, small form-factor compute, and data-efficient learning promises an attractive trade-off: modest hardware but large gains in perceptual robustness.

Looking carefully to the existing reviews, we found many surveys on sound source localization in the past three decades; however, to better illustrate new studies, we summarize the ones published after 2017 in Table \ref{tab:ssl_review_comparison}. A closer look at recent reviews indicates that no review has explored SSL in robotic platforms with a particular focus on deep-learning models. 

\begin{table*}[ht]
\centering
\caption{Comparative summary of sound–source-localisation (SSL) review papers.}
\label{tab:ssl_review_comparison}
\begin{tabular}{p{4.8cm} c c c p{5.5cm} c}
\toprule
\textbf{Review paper} &
\textbf{Ref.} &
\textbf{Year} &
\textbf{Time span} &
\textbf{Main focus} &
\textbf{Robotics} \\
\midrule
Localization of sound sources in robotics: A review &
\cite{rascon2017localization} &
2017 & Up to 2017 &
SSL in robotics; conventional techniques; SSL facets &
Yes \\[0.2em]

Localization of sound sources: A systematic review &
\cite{liaquat2021localization} &
2021 & 2011–2021 &
SSL methods; influencing factors; practical constraints &
No \\[0.2em]

Survey of sound source localization with deep learning methods &
\cite{grumiaux2022survey} &
2022 & 2011–2021 &
DL-based SSL techniques, architectures, datasets &
No \\[0.2em]

A review on sound source localization systems &
\cite{desai2022review} &
2022 & up to 2021 &
SSL systems in various array types and models &
Partial \\[0.2em]

Nonverbal sound in human–robot interaction: A systematic review &
\cite{zhang2023nonverbal} &
2023 & Up to April 2022 &
Non-verbal sound in HRI &
Yes \\[0.2em]

A survey of sound source localization and detection methods and their applications &
\cite{jekaterynczuk2023survey} &
2023 & Up to 2023 &
Classical + DL SSL methods and applications &
No \\[0.2em]

An overview of sound source localization based condition-monitoring robots &
\cite{lv2024overview} &
2024 & Up to 2024 &
SSL in condition-monitoring robots &
Yes \\[0.2em]

A review on recent advances in sound source localization techniques, challenges, and applications &
\cite{khan2025review} &
2025 & Up to 2025 &
General SSL system architectures and types &
No \\[0.2em]

\bottomrule
\end{tabular}
\end{table*}

In this regard, Rascon and Meza’s seminal survey \cite{rascon2017localization}, although could be the closest review to our review due to its robotic theme, predates the DL boom since it included papers before 2017. 
Liaghat et al. \cite{liaquat2021localization} had a broader focus to systematically review SSL works without focusing on specific applications (e.g. robotics), while reviewing the SSL methods between 2011 and 2021, overlooking the focus on DL models.
In contrast, Grumiaux et al. \cite{grumiaux2022survey} focus on new methods (from 2011 to 2021), especially in DL methods and their challenges. This review is very informative in the general domain of SSL and was very well cited, while it does not restrict its focus to robotics.
Review by Desai and Mehendale \cite{desai2022review}, surveyed the SSL works based on the number of microphones i.e. two microphones mimicking human auditory systems (binaural) and multiple microphones, and also based on the method (classical vs Convolutional neural networks based). They also provided very informative information in general SSL, such as the challenges of each SSL systems, while they didn't focus on robotics SSL.
Zhang et al. \cite{zhang2023nonverbal}, had a different focus such that it explored nonverbal sound in human-robot interaction by offering new taxonomies of function (perception vs creation) and form (vocables, mechanical sounds, etc.). While not strictly about SSL, it identified how sound contributes to robot perception and communication, and emphasized underexplored aspects like robot-generated sound and shared datasets. Their focus was novel and important for understanding how SSL integrates into broader auditory HRI.
Jekaterynczuk and Piotrowski \cite{jekaterynczuk2023survey} offered an extensive comparison of classical and AI-based SSL methods, with interesting classification by mic configuration, signal parameters, and neural architectures. They categorized the application of SSL works based on civil and military domains, and did not focus on robotics specifically.
Lv et al. \cite{lv2024overview} recently had a very interesting and specific review on SSL in condition monitoring robots (CMRs). They reviewed the diverse SSL techniques, including traditional and machine learning models. Their review was specifically narrowed to CMRs, and since SSL has not been extensively explored in there, they proposed a framework for future studies in this field. They also encourage future researchers in different condition monitoring tasks who use mobile robots to include SSL in addition to their existing monitoring systems, such as visual, infrared, etc.
The most recent review on SSL was carried out by Khan et al. \cite{khan2025review}, where they explored traditional SSL as well as machine learning models. They also categorized works based on different applications, such as industrial domains, medical science, and speech enhancement. 

This review fills the gap in other surveys by reviewing peer-reviewed literature from 2013 to 2024 in which different methods (especially Machine learning and deep learning) in SSL are applied to, or evaluated on, robotic platforms. It also explained the fundamental SSL facets and traditional SSL as well as DL methods. We also Explore the data, a pillar in deep learning models, together with different training strategies, then map SSL functions to concrete robotic tasks including service, social, search and rescue, and industrial applications and outline open challenges and future avenues.
The remainder of this review is organized accordingly such that  Section~\ref{sec:reviewmethod} details our literature-search and review. Section~\ref{sec:fundementals} revisits the acoustic foundations of SSL and the key environmental assumptions.  Section~\ref{sec:trad_dl} briefly outlines traditional SSL to contemporary deep-learning architectures. Section~\ref{sec:data+learning} discusses datasets and learning strategies underpinning DL-based SSL. Section~\ref{sec:apps} surveys how these techniques are deployed across different classes of robots and applications, and Section~\ref{sec:Challenges and Future Avenues} identifies research challenges and future avenues.

\section{Review Methodology}
\label{sec:reviewmethod}

Our goal is to paint a clear picture of how different methods, especially deep-learning approaches, in sound-source localization (SSL) are currently being used, tested, and deployed on robotic platforms. To achieve this, we carried out a targeted, multi-stage literature search prioritizing breadth of coverage over exhaustive enumeration.

\subsection{Literature Search Strategy}

Our search commenced by querying major engineering and robotics databases, including \textbf{IEEE Xplore}, \textbf{ACM Digital Library}, \textbf{ScienceDirect}, and \textbf{SpringerLink}. This was complemented by extensive use of \textbf{Google Scholar} to capture publications from emerging workshops and arXiv preprints that may have subsequently undergone peer review. We also manually inspected conference proceedings from key robotics venues such as ICRA, IROS, RSS, and IEEE RO-MAN, alongside principal audio forums like ICASSP, WASPAA, Interspeech, DCASE, and EUSIPCO.

The core of our search strategy involved combining three conceptual blocks using boolean operators. The primary block focused on sound localization terms, connected via "AND" to a block of robotics-related keywords. For instance, a common search string was:
\begin{center}
\small
\texttt{("sound source localization" OR "sound source localisation" OR
  "acoustic localization" OR "acoustic localisation" OR "sound source detection" OR "DOA estimation")}\\
\texttt{AND (robot* OR "mobile robot*" OR "service robot*" OR "industrial robot*" OR
  cobot* OR humanoid* OR "legged robot*" OR
  drone* OR UAV* OR quadrotor* OR multirotor*)}\\
% \texttt{AND ("deep learning" OR "neural network" OR CNN OR RNN OR Transformer)}\\
\end{center}
Variations such as \textit{``speaker localisation''}, \textit{``binaural CNN''}, or \textit{``SELD robot''} were iteratively added as citation chaining uncovered new terminology and relevant keywords.

% Our goal is to paint a clear picture of how different methods, especially deep-learning, in sound-source localization are currently being used, tested, and deployed on robotic platforms.  
% To that end we carried out a targeted, multi-stage literature search that prioritized breadth of coverage over exhaustive enumeration.

% \textbf{Sources consulted:}  
% We queried the major engineering and robotics databases—\textbf{IEEE Xplore}, \textbf{ACM Digital Library}, \textbf{ScienceDirect}, and \textbf{SpringerLink}—and complemented them with \textbf{Google Scholar} to catch publications in emerging workshops and arXiv preprints that were subsequently peer-reviewed.  
% Conference proceedings from the key robotics venues (ICRA, IROS, RSS, IEEE RO-MAN) and the principal audio forums (ICASSP, WASPAA, Interspeech, DCASE, EUSIPCO) were also inspected manually.

% \textbf{Keyword strategy:}  
% Search strings combined three concept blocks:

% \begin{center}
% \small
% \texttt{("sound source localization" OR "sound source localisation" OR "sound source detection" OR "acoustic localization" OR "acoustic localisation" OR "audio DOA")}\\
% \texttt{AND (robot* OR cobot* OR humanoid* OR mobile* OR service* OR industrial* OR drone* OR UAV*)}\\
% % \texttt{AND ("deep learning" OR "neural network" OR CNN OR RNN OR Transformer)}
% \end{center}

% Variations such as \textit{``speaker localisation''}, \textit{``binaural CNN''}, or \textit{``SELD robot''} were added iteratively when citation chaining uncovered new terminology.

\subsection{Inclusion and Exclusion Criteria}

To ensure the relevance and quality of the reviewed literature, a strict set of inclusion and exclusion criteria was applied during the screening process. Papers were primarily included if they were:
\begin{itemize}
    \item Peer-reviewed publications, encompassing journal articles, full conference papers, or workshop papers with archival proceedings.
    \item Published within the window of January 1, 2013, to May 1, 2025, capturing the significant surge of deep learning applications in robotics.
    \item Relevant to a robotic context, meaning the work either (i) evaluated SSL on a physical or simulated robot, or (ii) explicitly targeted a specific robotic use-case (e.g., service, industrial, aerial, field, or social Human-Robot Interaction).
    \item Written in English.
\end{itemize}
Conversely, studies were excluded if they were patents, magazine tutorials, or non-archival extended abstracts. Articles limited to headphone spatial audio, hearing aids, pure speech recognition, or architectural acoustics that lacked direct robotic relevance were also discarded.
% \subsection{Screening and Corpus Formation}

The screening procedure involved an initial scan of titles and abstracts to filter out papers clearly outside the defined scope. For the remaining records, a thorough full-text inspection was conducted. During this phase, particular attention was paid to the microphone configuration, the specific learning architecture employed, the evaluation protocol, and the presence of direct robotic experimentation or use-case targeting. Citation snowballing, both forward and backward, was applied to ensure that influential papers cited by the shortlisted works, or citing them, were not overlooked. This meticulous, iterative process ultimately converged on a corpus of \textbf{78 papers}, which form the basis of this comprehensive review.

\subsection{Publication Trends and Venues}

\begin{figure*}[htbp]
\centering
\includegraphics[width=0.7\textwidth]{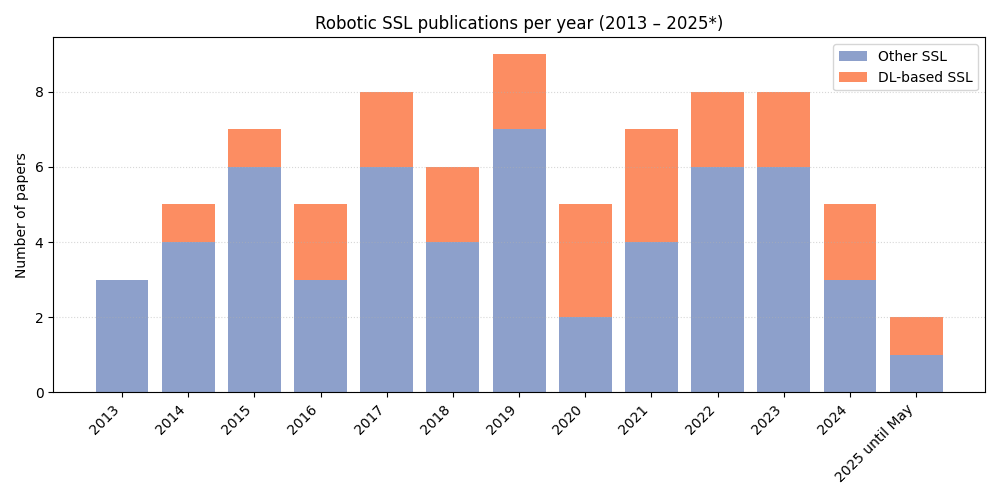} % Or .pdf, .jpg etc.
\caption{Number of papers published per year using SSL in robotics. The orange column represents the number of DL-based SSL, and the blue column shows SSL papers using other approacheses in robotics.}
\label{fig:pub_peryear} % This is the label you'd use in \ref{}
\end{figure*}

Figure \ref{fig:pub_peryear} illustrates the evolving landscape of SSL research in robotics through its annual publication trends. As observed, annual publications in SSL for robotics have consistently remained above five papers since 2014, reaching a peak of nine papers in 2019. The seemingly lower count for 2025, however, is attributed to our review's cut-off date of May 2025. Deep-learning approaches, notably absent before 2015, began to gain significant traction that year. Since 2020, they have consistently accounted for approximately one-third of all SSL-for-robotics publications, contributing a steady 2–3 papers annually and underscoring their growing prominence in the field.
% \textbf{Publication trends:}
% Through analysis of 78 papers, Figure \ref{fig:pub_peryear} is depicted. This Figure illustrates the evolving landscape of SSL research in robotics through its annual publication trends. As observed, annual publications in SSL for robotics have remained above five papers since 2014, reaching a peak of nine papers in 2019. The seemingly lower count for 2025, however, is attributed to our review's cut-off date of May 2025. Prior to 2016, deep-learning approaches were rare, but gained traction that year. Since 2020, they have consistently accounted for approximately one-third of all SSL-for-robotics publications, contributing a steady 2–3 papers annually and underscoring their growing prominence in the field.

% \begin{figure}[htbp]
%     \centering
%     \includegraphics[width=0.7\textwidth]{figs/Publications_Per_Year-SSL_in_Robotics.png} % Or .pdf, .jpg etc.
%     \caption{Number of papers published per year using SSL in robotics. The orange column represents the number of DL-based SSL, and the blue column shows SSL papers using other approacheses in robotics.}
%     \label{fig:pub_peryear} % This is the label you'd use in \ref{}
% \end{figure}

To guide future research dissemination in this dynamic field, Table \ref{tab:venue_breakdown_sorted} presents the most common publication venues (those with two or more papers) from our corpus of 78 reviewed articles. Among these 78 papers, 36 were published as conference papers, 1 as a PhD thesis \cite{he2021deep}, and 41 as journal papers. The table highlights the significant role of prominent conferences in SSL-for-robotics research. Notably, the two most prestigious robotics conferences, IROS (IEEE/RSJ International Conference on Intelligent Robots and Systems) and ICRA (IEEE International Conference on Robotics and Automation), show substantial contributions, with 11 and 3 papers respectively. ICASSP (International Conference on Acoustics, Speech, and Signal Processing), a key conference venue in acoustic signal processing, also features prominently with 4 papers in our review. In terms of journals, Table \ref{tab:venue_breakdown_sorted} indicates that only the IEEE Sensors Journal has contributed more than two papers to our review corpus; other journals are represented by fewer articles each.

% \textbf{Common publication venues:}
% To guide future research dissemination in this field, Table \ref{tab:venue_breakdown_sorted} presents the most common publication venues (those with two or more papers) from our corpus of 78 reviewed articles. 
% Among the 78 papers, 36 were published as conference papers, 1 as a PhD thesis \cite{he2021deep}, and 41 as journal papers.
% The table highlights the significant role of prominent conferences in SSL-for-robotics research. Notably, the two most prestigious robotics conferences, IROS (IEEE/RSJ International Conference on Intelligent Robots and Systems) and ICRA (IEEE International Conference on Robotics and Automation), show substantial contributions, with 11 and 3 papers respectively. ICASSP (International Conference on Acoustics, Speech, and Signal Processing), a key conference venue in acoustic signal processing, also features prominently with 4 papers in our review. In terms of journals, Table \ref{tab:venue_breakdown_sorted} indicates that only the IEEE Sensors Journal has contributed more than two papers to our review corpus; other journals are represented by fewer articles each.
% % indicates that only five journals have contributed more than one paper to our selected corpus.
% % The presence of not is because SSL in robotics can cover a wide spectrum of journals, from journals with a robotic focus to acoustic or some with a specific domain. 

\begin{table*}[ht]
    \centering
    \caption{Most published venues among the papers in robotic SSL in our review. In the third column, "C" and "J" stand for conference and journal.}
    \label{tab:venue_breakdown_sorted}
    \begin{tabular}{@{}lccp{5cm}@{}}
        \toprule
        \textbf{Venue} & \textbf{No. of papers} & \textbf{C/J} & \textbf{Reference} \\ \midrule
        IROS
            & 11 & C & \cite{youssef2013learning,nakamura2013real,ohata2014improvement,grondin2015time,nakamura2015interactive,li2016reverberant,nakadai2017development,strauss2018dregon,wang2019audio,michaud20203d,sewtz2020robust} \\[2pt]
        ICASSP
            & 4 & C & \cite{tourbabin2015enhanced,takeda2016sound,takeda2017unsupervised,ferguson2018sound} \\
        ICRA
            & 3 & C & \cite{grondin2016noise,he2018deep,an2018reflection} \\

        \textit{IEEE Sensors Journal}
            & 3 & J & \cite{latif2015sound,song2020automatic,clayton2023embedded} \\[2pt]

        \textit{Robotics and Autonomous Systems}
            & 2 & J & \cite{grondin2019lightweight,rascon2017localization} \\
        \textit{Drones}
            & 2 & J & \cite{go2021acoustic,yamada2023placement} \\
        \textit{Applied Sciences}
            & 2 & J & \cite{skoczylas2021belt,shi2023audio} \\
        \textit{IEEE Transaction on Instrumentation}\\ \textit{
         and Measurement}
            & 2 & J & \cite{keyrouz2014advanced,wang2023multiple} \\[2pt]
        \textit{IEEE$/$ACM Transactions on Audio, Speech,}\\ \textit{and Language Processing}
            & 2 & J & \cite{tourbabin2014theoretical,manamperi2022drone} \\[2pt]
        ICDL
            & 2 & C & \cite{gonzalez2021self,gamboa2022real} \\ \bottomrule
    \end{tabular}

\end{table*}

\section{SSL foundementals}
\label{sec:fundementals}

Sound source localization (SSL) is the process of determining the spatial location of one or more sound sources based on measurements from acoustic sensors. This section provides an overview of the fundamental principles and terminology that form the foundation of SSL in robotics applications.

Sound propagates through air as pressure waves, traveling at approximately 343 meters per second under standard conditions. When these waves encounter microphones or other acoustic sensors, they are converted into electrical signals that can be processed to extract spatial information. The core principle underlying SSL is that sound reaches
different spatial positions at different times and with different intensities, creating patterns that can be analyzed to determine the source's location. In the context of robotics, SSL typically involves estimating Direction of Arrival (DOA) and distance or complete position. DOA is defined as an angle or vector pointing toward the sound source, often expressed in terms of azimuth (horizontal angle) and elevation (vertical angle). Distance is the range between the sound source and the receiver, and position is the complete three-dimensional coordinates of the sound source in space. According to the objectives of each study, SSL tasks can include estimating DOA, distance, or position of the sound source.

As noted by Rascon and Meza \cite{rascon2017localization}, the majority of SSL systems in robotics focus primarily on DOA estimation, as distance estimation presents additional challenges and is often less critical for many applications. 
% The authors state that "SSL in real-life scenarios needs to take into account that more than one sound source might be active in the environment," highlighting the complexity of the task in practical settings.
One challenge in SSL is taking into consideration that the sound source (or sources) might be active, a source is active when emitting sound, or inactive during the localization task. Therefore, considering a source (or sources) as always active might be unrealistic in a practical setting. To deal with this challenge, as extensively pointed out in a survey by Grumiaux et al. \cite{grumiaux2022survey}, the source activity detection can be done either before (as a separate task) or simultaneously within SSL task, for example a neural network predicts both location and activity of a sound source \cite{yalta2017sound}) .
It is important to note that sound source localization is broader than sound event detection (SED), where in SSL, the location of sound sources is obtained, as well as the detection of the presence of sound. Therefore, in many studies, SSL is referred to as sound event localization and detection (SELD) \cite{grumiaux2022survey}. It is also worthwhile mentioning that sound source separation (in the presence of multiple active sounds) is another task that has to be done in SSL when dealing with more than one active sound source \cite{chen2022multiple}. Therefore, some studies are focusing on sound source counting as well as localization (such as \cite{tian2020multiple}), while many assume we have prior knowledge of sound source numbers in the scene (e.g. \cite{bohlender2021exploiting}).

\subsection{Acoustic Signal Propagation}

Understanding acoustic signal propagation is essential for developing effective SSL systems. In ideal free-field conditions, sound waves propagate spherically from a point source, with amplitude decreasing proportionally to the distance from the source. However, real-world environments introduce several complexities: 

\textbf{Reverberation:} Sound reflections from surfaces create multiple paths between the source and receiver, complicating the localization process.

\textbf{Diffraction:} This phenomenon happens where sound waves bend or spread around obstacles or through openings (like doorways or windows). It allows sound to be heard even when there's no direct line of sight to the source.

\textbf{Refraction:} Is the bending of sound waves as they pass from one medium into another, or as they travel through a medium where the speed of sound changes gradually. The speed of sound can vary due to changes in temperature, wind, or medium Density.

\textbf{Background Noise (external noise) and Sensor Imperfection (internal noise):} sounds from ambient, or noise generated from the robot operation itself \cite{li2011sound}, can mask or interfere with the target source; this noise can be referred to as external noise. Also, due to the imperfection of the receiver system, the recorded sound can be deviated from what it should be correctly recorded because of microphones or the audio acquisition system (e.g., analog-to-digital converters). This noise is inherent to the sensing hardware and its associated electronics.

These phenomena significantly impact the performance of SSL systems and have driven the development of increasingly sophisticated algorithms to address these challenges.
To deal with these challenges, some studies considered some assumptions to simplify the SSL. For example, some early studies considered that there is no reverberation in the environment; this setting is called "anechoic". Despite not being realistic in most applications, the anechoic setting has been assumed in many SSL works \cite{grumiaux2022survey,el2000neural,ishfaque2022real}. 
To deal with noise, some studies used denoising techniques to overcome noise (e.g. \cite{athanasopoulos2012effect}), while some considered the effect of noise levels, as defined by SNR (Signal-to-Noise ratio), in the localization of sound sources \cite{subramanian2022deep,goli2023deep,wu2024sound,jalayer2024convlstm}.

\subsection{Coordinate Systems and Terminology}

SSL systems typically employ either Cartesian or spherical coordinates in both 2D and 3D localization. In Cartesian Coordinates, the positions of sound sources are obtained with respect to the X, Y, and Z axes in 3D (X and Y in 2D). In Spherical coordinates, the location of sound sources are determined in terms of radius (distance), azimuth, and elevation angles. In 2D localization, many studies focused on azimuth and distance or azimuth (horizontal localization \cite{risoud2018sound}) or azimuth and elevation (directional localization). Also some works only restricted their objective to azimuth angle (1D localization) relative to the microphone array position, and in some cases, they do localization grid by dividing the 360 degrees azimuth angle into grid space, such as 8 sections \cite{hirvonen2015classification} or narrower grid space (e.g. 360 sections \cite{xiao2015learning}). This trick aims to perform SSL as a classification task in machine learning, where each class is devoted to a specific subregion for estimating sound sources \cite{grumiaux2022survey}.

In robotics applications, the choice of coordinate system often depends on the specific requirements of the task and the configuration of the robot's sensors. 
Also, it is worthwhile mentioning that studies focusing on mobile robots might benefit from using the robot-centric coordinate (centered on the microphones mounted on the robot \cite{geng2008sound,liu2018sound}). On the other hand, a fixed and static coordinate system defined by the environment itself (e.g., a corner of a room, a fixed marker) may be preferred for stationary applications, e.g., industrial robotic arms that have fixed locations in the workplace.
% For example, one study might be more into the detection of azimuth angle for rotating the robot head towards a sound source (), therefore it is easier for them to just focus on angular coordinates than Cartesian.
% mobile robots, egocentric coordinate systems (centered on the robot) are commonly used, while fixed coordinate systems may be preferred for stationary applications.

\subsection{Microphone Numbers and Array Configurations} 

\begin{figure}
    
    \centering
        \centering
        \includegraphics[width=\linewidth]{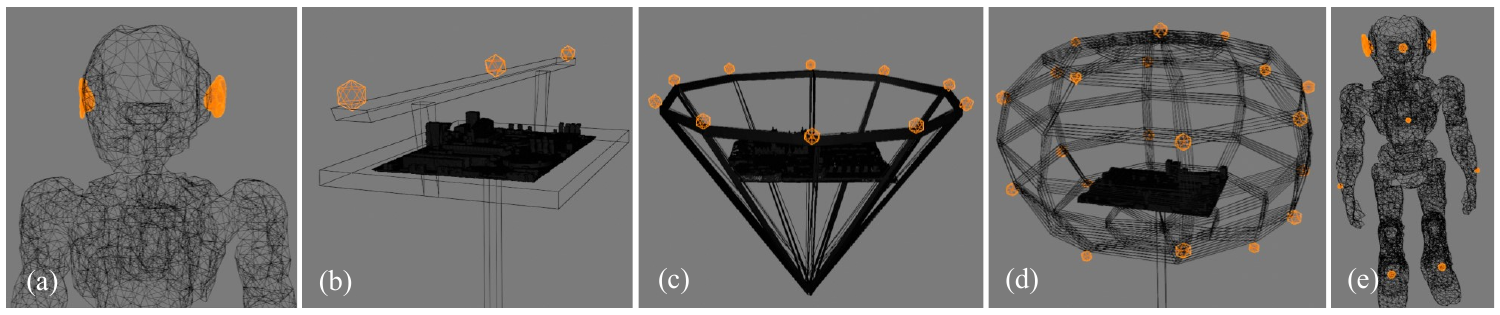}
        \caption{Binaural microphone used in \cite{keyrouz2014advanced}}
        \label{fig:binaural}
    % Second image: Hand Detection by Keypoints
    % \begin{subfigure}{0.3\textwidth}
    %     \centering
    %     \includegraphics[width=\linewidth]{figs/mics linear.png}
    %     \caption{Linear microphone used in \cite{sewtz2020robust}.}
    %     \label{fig:linear}
    % \end{subfigure}
    % % Third image: Hand Segmentation
    % \begin{subfigure}{0.3\textwidth}
    %     \centering
    %     \includegraphics[width=\linewidth]{figs/mics circular.png}
    %     \caption{Circular microphone used in \cite{faraji2019sound}.}
    %     \label{fig:circular}
    % \end{subfigure}
    % \begin{subfigure}{0.35\textwidth}
    %     \centering
    %     \includegraphics[width=\linewidth]{figs/mics spherical2.png}
    %     \caption{Spherical microphone used in \cite{sasaki2012spherical}.}
    %     \label{fig:spherical}
    % \end{subfigure}
    % \begin{subfigure}{0.50\textwidth}
    %     \centering
    %     \includegraphics[width=\linewidth]{figs/mics distributed.png}
    %     \caption{microphone placed in different NAO robot parts in \cite{tourbabin2015enhanced}.}
    %     \label{fig:arbitary}
    % \end{subfigure}
    \caption{Different microphone array configurations used in SSL robotics studies: (a) Binaural microphone array, (b) Linear array, (c) Circular array, (d) Spherical array, (e) microphones distributed in robot parts. }
 \label{fig:micgeometry}
 
\end{figure}

The arrangement of microphones plays a crucial role in SSL performance.
Before explaining the different configurations, it is important to describe an open challenge in this regard. As a general rule, a large number of microphones in SSL leads to high accuracy in localization \cite{lee2021deep}. However, including more microphones can cause higher computing, higher cost, and consequently higher latency in localization (not reaching real-time localization). Also, the variation in microphone number and arrangements adds other designing hyperparameters (how many microphones? which arrangement is better?). These additional hyperparameters in each study results in not having a reference SSL design system as a benchmark. On the other hand, because of some constraints in each study, e.g., different objectives, different financial and computational budgets, and design constraints, it is not expected that each study follow the same microphone design. Various microphone configuration have been used in SSL studies in robotics, as shown in Figure \ref{fig:micgeometry}. The variety in microphone array configurations can be categorized into the following configurations:

\textbf{Binaural Arrays:} Mimicking human hearing with two microphones \cite{acosta2023remote}, often placed on a robot's head or within a dummy head structure \cite{keyrouz2014advanced,deleforge2015acoustic,gala2019realtime}, as represented in Figure \ref{fig:micgeometry}a. These are particularly common in humanoid robots, or simple mobile robots \cite{baxendale2019feed,gala2023moving}, and offer natural spatial cues but may have limited resolution. The robot's head (or dummy head) can significantly impact sound waves through diffraction. This causes sound waves to bend around the head, leading to fluctuations in the Time Difference of Arrival (TDoA), especially for sounds traveling around the front and back. The front-back ambiguity challenge in binaural SSL for robots, which means the robot can struggle to differentiate between a sound coming from "ahead" or "behind" without additional cues or head movements. Binaural arrays are excellent at localizing sounds in the horizontal plane (azimuth) but offer very limited resolution for elevation (vertical direction). It's hard for them to tell if a sound is coming from above or below. These challenges lead to struggling with multiple simultaneous sound sources without using advanced separation techniques.

\textbf{Linear Arrays:} Microphones arranged in a straight line (see Figure \ref{fig:micgeometry}b), providing good resolution in one dimension but suffering from front-back ambiguity and limited elevation estimation. The microphones are typically equally spaced, but non-uniform spacing can also be used to optimize performance \cite{sewtz2020robust}. In addition to simplicity of design and implementation, this configuration is easier to calibrate than complex arrangements. A linear array mounted on the front or top of a robot can effectively localize sounds in the horizontal plane ahead of the robot \cite{mumolo2003algorithms,nguyen2017long}. This is useful for directional voice commands or detecting sounds from the front.

\textbf{Circular Arrays:} As shown in Figure \ref{fig:micgeometry}c, Microphones arranged in a circle, offering 360-degree coverage in the horizontal plane and eliminating front-back ambiguity. Mounted on the "head" or "torso" of social robots to perceive sounds from any direction around them, crucial for engaging with multiple people in a room \cite{tamai2004circular,choi2003speech,grondin2019lightweight}.

\textbf{Spherical Arrays:} Microphones distributed over a spherical surface, as can be seen in Figure \ref{fig:micgeometry}d, providing full three-dimensional coverage and the ability to decompose the sound field into
spherical harmonics. The spherical arrays can enable a robot \cite{sasaki2012spherical} or a drone \cite{nakadai2017development,yamada2020sound,yamada2023placement} to precisely localize and track multiple speakers in 3D space, understanding who is speaking from where in complex social settings.

\textbf{Arbitrary geometries:} This refers to microphone configurations where the individual microphone elements are positioned without adherence to the particular geometric arrays. The placement might be dictated by factors external to optimal acoustic design, such as the physical constraints of a platform, aesthetic considerations, the repurposing of existing sensors, or opportunistic placement in an environment. In robotics, microphones could be placed in different parts of the body as can be seen in Figure \ref{fig:micgeometry}e, e.g., the head, torso, and limbs of a humanoid robot \cite{tourbabin2014theoretical,tourbabin2015enhanced} or distributed inside the room \cite{jin2020rnn,jalayer2024convlstm}. The irregular microphone placements have also been implemented in non-humanoid such as a hose-shaped robot \cite{bando2013posture,mae2017sound}

% \begin{figure*}
    
%     \centering
%     \begin{subfigure}{0.35\textwidth}
%         \centering
%         \includegraphics[width=\linewidth]{figs/mics binaural.png}
%         \caption{Binaural microphone used in \cite{keyrouz2014advanced}}
%         \label{fig:binaural}
%     \end{subfigure}
%     % Second image: Hand Detection by Keypoints
%     \begin{subfigure}{0.3\textwidth}
%         \centering
%         \includegraphics[width=\linewidth]{figs/mics linear.png}
%         \caption{Linear microphone used in \cite{sewtz2020robust}.}
%         \label{fig:linear}
%     \end{subfigure}
%     % Third image: Hand Segmentation
%     \begin{subfigure}{0.3\textwidth}
%         \centering
%         \includegraphics[width=\linewidth]{figs/mics circular.png}
%         \caption{Circular microphone used in \cite{faraji2019sound}.}
%         \label{fig:circular}
%     \end{subfigure}
%     \begin{subfigure}{0.35\textwidth}
%         \centering
%         \includegraphics[width=\linewidth]{figs/mics spherical2.png}
%         \caption{Spherical microphone used in \cite{sasaki2012spherical}.}
%         \label{fig:spherical}
%     \end{subfigure}
%     \begin{subfigure}{0.50\textwidth}
%         \centering
%         \includegraphics[width=\linewidth]{figs/mics distributed.png}
%         \caption{microphone placed in different NAO robot parts in \cite{tourbabin2015enhanced}.}
%         \label{fig:arbitary}
%     \end{subfigure}
%     \caption{Different microphone array configurations used in SSL robotics studies.}
%  \label{fig:micgeometry}
 
% \end{figure}

% \textbf{Distributed Arrays:} Microphones placed at various locations throughout an environment, potentially offering improved coverage and triangulation capabilities.

The choice of the number of microphones and array configuration depends on the intention of researchers in each study. In robotics, many researchers want to mimic the human auditory system and chose binaural arrays. Also, it is noted that binaural is the most used configuration in SSL in robotics \cite{rascon2017localization}. Interestingly, some researchers made some advancements by designing external pinnae for robots to mimic the human auditory system more than just putting two microphones on the head of the robot \cite{kim2013improvement,kumon2011active,keyrouz2014advanced}. Regarding the number of microphones, the number of microphones varied a lot \cite{rascon2017localization}. It includes some SSL studies having a single microphone \cite{murray2011neural,zhang2001grounded,ishfaque2022real}, and some having a very dense microphone array (even 64 microphones \cite{sasaki2012spherical}).

% factors such as the robot's physical constraints, the required spatial resolution, and the specific application context. As noted 
% As noted by Grumiaux et al. (2022), recent advances in deep learning approaches have shown varying degrees of effectiveness across different array configurations, with some neural architectures being specifically designed to exploit the geometric properties of particular arrays.

\section{From Traditional SSL to DL Models}
\label{sec:trad_dl}

As robots are increasingly deployed in real-world settings, ranging from domestic assistants to industrial inspectors, the demands placed on SSL systems have grown considerably. These systems must now contend with complex, noisy, and reverberant environments, while maintaining low latency, minimal power consumption, and high spatial accuracy.
Over the past decades, SSL methodologies have evolved in response to these challenges, beginning with analytically grounded, signal-processing techniques that exploit physical properties of sound propagation and microphone arrays. While traditional methods such as Time Difference of Arrival (TDOA), beamforming, and subspace analysis laid a robust foundation, their performance often degrades under non-ideal conditions and hardware constraints typical of mobile and embedded robotic systems. These limitations, in turn, have catalyzed a shift toward data-driven and learning-based paradigms, promising greater adaptability and robustness in real-world applications.

\subsection{Traditional Sound Source Localization Methods}

Before neural networks entered the scene, robot audition drew almost exclusively on classical array-signal-processing.  Although most of today’s deep models replace, or embed, those analytical blocks, an understanding of their logic remains essential because the same acoustic constraints (microphone geometry, reverberation, noise, and multipath) still apply.  Below we revisit the four pillars that have shaped the field, i.e., \emph{time-delay estimation}, \emph{beamforming}, \emph{Steered-response power (SRP)}, and \emph{subspace analysis}.  Throughout we highlight why each method proved attractive for robots and where its weaknesses motivated the subsequent shift to learning-based pipelines.

\subsubsection{Time-difference of arrival (TDOA).}  
% TDOA measures how long a wavefront takes to travel between pairs of microphones and converts those delays into hyperbolic loci whose intersections reveal the source position.  In practice delays are obtained by maximising a cross-correlation; the phase-transform variant GCC-PHAT became the community’s work-horse because it attenuates spectral coloration and thus survives moderate reverberation \cite{rascon2017localization,grumiaux2022survey}.  On small mobile robots the algorithm’s appeal is twofold: it runs in real time on embedded CPUs and it works with very few sensors.  Yet coarse sampling, short baselines, front–back ambiguity in linear arrays, and interference between simultaneous talkers limit its accuracy.
Time Difference of Arrival (TDOA) is one of the foundational techniques in sound source localization (SSL), widely applied due to its conceptual simplicity and computational efficiency. The method relies on the fundamental observation that a sound wave will arrive at different microphones in an array at slightly different times, depending on the location of the source relative to the array \cite{grumiaux2022survey}. By measuring these inter-microphone time delays, one can infer the direction or position of the sound source \cite{xu2012high}.
At the heart of TDOA-based SSL is the cross-correlation of audio signals recorded at different microphones. This process identifies the time lag at which the similarity between two signals is maximized, corresponding to the delay caused by the sound wave’s path difference. To enhance the performance of this basic cross-correlation, especially in noisy or reverberant environments, researchers often employ Generalized Cross-Correlation (GCC), which introduces a frequency-dependent weighting to the correlation process \cite{knapp2003generalized}.
Among the variants of GCC, the Phase Transform (PHAT) weighting, commonly known as GCC-PHAT, has proven particularly effective in robotics contexts \cite{park2007performance,wang2021gcc}. By emphasizing phase information and attenuating amplitude components, GCC-PHAT improves robustness to reverberation and environmental noise. 
% As Rascon and Meza (2017) observed, GCC-PHAT has been substantially advanced by the robotics community and has become a de facto standard in mobile and service robot auditory systems.
Once the time delays are estimated, spatial localization is achieved through hyperbolic positioning. Each pairwise time difference corresponds to a hyperbolic curve (in 2D) or a hyperboloid surface (in 3D) upon which the sound source must lie. The intersection of these geometric constraints from multiple microphone pairs yields an estimate of the source’s location.

Despite its popularity, the TDOA method is not without limitations. One major challenge is its sensitivity to noise and reverberation \cite{lombard2010tdoa,huang2014tdoa}, particularly in enclosed environments with reflective surfaces. Although GCC-PHAT offers partial mitigation, its effectiveness diminishes in highly cluttered acoustic spaces. Additionally, the spatial resolution of TDOA-based systems is inherently limited by the sampling rate of the microphones and the inter-microphone distances, which constrain the granularity of detectable delays.
Another well-documented drawback is the so-called front–back ambiguity, which arises primarily in linear microphone arrays. In such configurations, the system may struggle to distinguish whether a sound originates from in front of or behind the array due to symmetric time-delay profiles. Moreover, traditional TDOA techniques face difficulties in multi-source environments \cite{scheuing2008disambiguation}, where overlapping sound signals can interfere with accurate time-delay estimation, leading to degraded performance or incorrect localization.
Nonetheless, TDOA remains a cornerstone of SSL research and application, especially in scenarios where computational simplicity and real-time performance are prioritized. Its principles have also served as a basis for hybrid systems that integrate TDOA with more advanced signal processing or learning-based techniques, expanding its utility in modern robotic auditory systems \cite{kim2013improved,kim2015improved,grondin2015time,grondin2016noise,chen2019sound,chen2020efficient,xu2013sound,alameda2014geometric,an2018reflection,strauss2018dregon,chen2019sound}.

\subsubsection{Beamforming}  

Beamforming is a spatial signal processing technique that plays a dual role in auditory systems: it not only facilitates sound source localization (SSL) but also enhances the quality of the captured audio by amplifying signals from a desired direction while attenuating noise from others. In robotic contexts, this capability is invaluable for tasks such as speech recognition, situational awareness, and interaction in acoustically complex environments \cite{valin2007robust,luzanto2024effective,kagami20092d}.
At its core, beamforming involves the constructive combination of sound signals from multiple microphones, typically after delaying them to align with a hypothesized source direction. When the hypothesized direction matches the true direction of arrival (DOA), the signals reinforce, resulting in a high beam response. The simplest implementation of this principle is the Delay-and-Sum (DS) beamformer, which sums delayed microphone signals to estimate the DOA. More sophisticated approaches like the Minimum Variance Distortionless Response (MVDR) beamformer, also known as Capon’s beamformer \cite{capon2005high}, improve robustness by minimizing output power from all directions except the desired one. The Linearly Constrained Minimum Variance (LCMV) beamformer allows multiple spatial constraints, making it suitable for multi-target environments \cite{dmochowski2008linearly}. Adaptive beamforming methods extend this paradigm by dynamically adjusting parameters in response to changes in the environment, enhancing resilience to noise and interference.

% Recent research has focused on enhancing beamforming to meet the demands of robotic applications in challenging conditions.
In the survey by Lv et al. \cite{lv2024overview}, some promising developments in beamforming were explained that could be implemented as beamforming-based SSL in robotics. For example, a novel Beamforming technique by Yang et al. \cite{yang2022improved} tailored for far-field, large-scale environments significantly improved localization accuracy across multiple sound sources. In low signal-to-noise ratio (SNR) conditions, Liu et al. \cite{liu2023minimum} introduced a novel MVDR-based beamforming (MVDR-HBT) algorithm, which leverages statistical signal properties to boost robustness and precision. Zhang et al. \cite{zhang2023localization} extended beamforming to periodic, steady-state sound sources by developing a high-resolution cyclic beamforming method that supports both localization and fault diagnosis in machinery.
While beamforming offers considerable advantages, it is not without its challenges. Traditional beamforming techniques tend to suffer from poor dynamic range and limited real-time performance, particularly in large environments or when precise temporal resolution is required. Moreover, accurate array calibration remains a critical prerequisite; small discrepancies in microphone positioning or gain can lead to substantial localization errors. The computational complexity of adaptive beamforming in real-time applications can also place significant demands on embedded processors typical of mobile robotic platforms and UAVs \cite{faraji2019sound,go2021acoustic}.
Nonetheless, beamforming remains a vital technique in SSL which could be used in many robotic applications. Its capacity for both directional enhancement and localization makes it particularly attractive for systems that require perceptual robustness in dynamic or noisy environments.

\subsubsection{Steered-response power (SRP).}  
% SRP methods take a search-and-verify stance: a virtual beamformer is steered toward every candidate direction on a grid and the output energy is logged; the peak identifies the source \cite{khan2025review}.  SRP-PHAT inherits GCC robustness while exploiting the full array aperture, which explains its use in circular eight-mic robots powered by HARK.  The price to pay is computational: a fine angular lattice entails thousands of delay-and-sum evaluations per frame, and diffuse noise broadens the SRP peak, masking weaker sources.

Steered Response Power (SRP) methods constitute a prominent category of traditional SSL techniques. These methods operate on the principle of beamforming: the microphone array is virtually “steered” in multiple candidate directions, and the response power, essentially the energy of the summed signals after delay alignment, is computed for each direction. The underlying assumption is that when the beamformer is steered toward the true source direction, the accumulated energy or response power will be maximized.
Among the various SRP techniques, the SRP-PHAT (Phase Transform) algorithm is widely recognized for its robustness and practical effectiveness. It incorporates the phase transform weighting strategy from GCC-PHAT into the SRP framework, attenuating amplitude information and emphasizing phase cues to improve reliability in reverberant environments \cite{dibiase2001robust}.

One of the most valuable attributes of SRP methods (particularly SRP-PHAT) is their resilience in acoustically challenging settings. Unlike traditional cross-correlation techniques, SRP methods perform well even in the presence of significant reverberation, making them suitable for indoor or cluttered robotic applications. 
Another advantage is their capacity for multi-source localization. By analyzing the response power map across a spatial grid, systems can identify multiple peaks, each corresponding to a potential sound source. This ability makes SRP methods appealing in scenarios where robots must interact with multiple humans or monitor several machines simultaneously.
However, SRP approaches are not without challenges. Chief among them is computational complexity. Evaluating the steered response across a dense spatial grid requires substantial processing, which can be burdensome for real-time systems or power-constrained platforms. Moreover, the spatial resolution of the localization is directly tied to the granularity of the search grid. Finer grids improve accuracy but exacerbate computational demands.
Discretization also introduces errors, particularly when the true source lies between predefined grid points. Additionally, while SRP-PHAT improves robustness to reverberation, performance may still degrade in noisy environments or when multiple sources overlap in time and frequency, leading to ambiguities in peak detection.
Despite these limitations, SRP-based methods, especially when paired with optimization strategies or hierarchical search techniques, remain a powerful tool in the SSL toolbox. They bridge the gap between theoretical accuracy and real-world applicability and continue to be refined for emerging robotic use cases that demand high reliability in dynamic, reverberant, and multi-source auditory scenes \cite{gamboa2022real,yook2015fast,li2016reverberant,grondin2019lightweight,wang2019audio,michaud20203d}.

\subsubsection{Subspace analysis}  

Subspace-based methods form one of the most analytically powerful approaches in the domain of sound source localization (SSL). These techniques, including the well-known Multiple Signal Classification (MUSIC) and Estimation of Signal Parameters via Rotational Invariance Techniques (ESPRIT), operate by analyzing the eigenspectrum of the spatial covariance matrix derived from multichannel microphone signals. Their strength lies in the decomposition of this matrix into orthogonal signal and noise subspaces, which enables the extraction of direction-of-arrival (DOA) information with high angular precision.
The MUSIC algorithm remains a prominent representative of this class. It estimates the covariance matrix of the received signals and then applies eigendecomposition to distinguish the dominant signal subspace from the residual noise subspace. A spatial pseudo-spectrum is then generated by scanning over candidate directions; peaks in this spectrum correspond to estimated DOAs, exploiting the principle that the array steering vectors of true sources are orthogonal to the noise subspace \cite{ishi2009evaluation}.

One of the significant limitations of traditional MUSIC is its dependency on accurate and noise-free estimation of the covariance matrix. In real-world scenarios with limited data or low signal-to-noise ratios (SNR), these estimations can become unreliable. To address this, \citet{zhang2022efficient} proposed the use of a non-zero delay sample covariance matrix (SCM) combined with pre-projection techniques to filter out noise and improve signal subspace estimation. Similarly, \citet{weng2023doa} introduced the SHD-BMUSIC algorithm, which operates in the spherical harmonic domain and integrates wideband beamforming to enhance source discrimination, particularly in scenarios involving closely spaced or multiple sources.
Despite their theoretical elegance and high spatial resolution, subspace methods are not without drawbacks. They are sensitive to coherent or correlated sound sources, a common condition in reverberant environments, which can collapse the signal subspace and compromise accuracy. Moreover, these methods demand significant computational resources \cite{wang2023multiple}, primarily due to eigendecomposition and exhaustive spatial scanning. They also typically require a substantial number of temporal snapshots to yield stable covariance estimates, reducing their responsiveness in rapidly changing acoustic scenes.
Nonetheless, subspace methods remain an essential pillar of SSL research. Their precision in controlled environments and potential for multi-source resolution make them attractive for applications in collaborative mobile ground robotics \cite{suzuki2017development,chen2022broadband,chen2022large,shi2023audio} and UAVs \cite{ohata2014improvement,hoshiba2017design,nakadai2017development,manamperi2022drone,yamada2023placement,azrad2024performance} as well as humanoid audition \cite{nakamura2009intelligent,ishi2009evaluation,nakamura2013real,narang2014auditory,sewtz2020robust,asano2015sound}. Ongoing research aims to mitigate their limitations through techniques such as subspace smoothing, sparse array design, and integration with learning-based models, enabling their gradual transition from theoretical benchmarks to practical solutions in robotic auditory systems.

Table \ref{tab:traditional_ssl_robots} lists the information regarding the traditional SSL works in robotics that used the four typical localization families. As the table shows, a considerable number of papers in the last decade still rely on these classical SSL approaches (especially subspace analysis using MUSIC) in robotics. Interestingly, the vast majority of these studies target one stationary sound source in their localization experiments. This emphasis reflects an inherent limitation of traditional SSL techniques in tracking multiple or moving emitters with high accuracy.

\begin{table*}[htbp]
\footnotesize            % ↓ shrink text so the table fits the page
\setlength{\tabcolsep}{5pt}   % ↓ tighter column spacing
\centering
\caption{Studies on robotic platforms in our review that used the typical traditional SSL approaches (i.e., TDOA, beamforming, SRP, and subspace analysis).  “S/M’’ denotes \textbf{S}tatic / \textbf{M}oving sound sources.}
\label{tab:traditional_ssl_robots}
\resizebox{\textwidth}{!}{%
\begin{tabular}{ccccccc}
\toprule
\textbf{Paper} & \textbf{Method} & \textbf{Year} & \textbf{Robot Type} & \textbf{Max.~active sources} & \textbf{S/M} & \textbf{No. of Microphones (geometry)} \\
\midrule
\cite{xu2013sound}            & TDOA                 & 2013 & Mobile ground      & 1  & S       & 4 (tetrahedral) \\
\cite{kim2013improved}        & TDOA                 & 2013 & Humanoid           & 1  & S       & 2 (binaural)    \\
\cite{alameda2014geometric}   & TDOA                 & 2014 & General            & 1  & S       & 4              \\
\cite{kim2015improved}        & TDOA                 & 2015 & Humanoid           & 1  & S       & 2 (binaural)    \\
\cite{grondin2015time}        & TDOA                 & 2015 & Mobile ground      & 1  & S       & 2              \\
\cite{grondin2016noise}       & TDOA                 & 2016 & Mobile ground      & 1  & S       & 2              \\
\cite{an2018reflection}       & TDOA                 & 2018 & Mobile ground      & 1  & S\,+\,M & 8              \\
\cite{strauss2018dregon}      & TDOA                 & 2018 & UAV                & 1  & S       & 8              \\
\cite{chen2019sound}          & TDOA                 & 2019 & Mobile ground      & 1  & S       & 5 (pyramid)    \\
\cite{chen2020efficient}      & TDOA                 & 2020 & Humanoid           & 1  & M       & 2 (binaural)    \\
\cite{faraji2019sound}        & Beamforming          & 2019 & UAV                & 1  & M       & 8 (circular)    \\
\cite{go2021acoustic}         & Beamforming          & 2021 & UAV                & 1  & S       & 32 (circular)   \\
\cite{yook2015fast}           & SRP                  & 2015 & General            & 1  & S       & 16 (cylindrical)\\
\cite{li2016reverberant}      & SRP                  & 2016 & Humanoid           & 1  & S       & 4              \\
\cite{grondin2019lightweight} & SRP                  & 2019 & Mobile ground      & 4 & S\,+\,M & 8 / 16 (circular and closed cubic) \\
\cite{wang2019audio}          & SRP                  & 2019 & UAV                & 2  & S       & 8 (circular)    \\
\cite{michaud20203d}          & SRP                  & 2020 & Mobile ground      & 1  & S\,+\,M & 16             \\
\cite{gamboa2022real}         & SRP                  & 2022 & Mobile ground      & 1  & S       & 3              \\
\cite{azrad2024performance}   & SRP\,/ subspace analysis     & 2024 & UAV                & 1  & S       & 6              \\
\cite{nakamura2013real}       & Subspace analysis    & 2013 & Humanoid           & 1  & M       & 16 (circular)   \\
\cite{ohata2014improvement}   & Subspace analysis    & 2014 & UAV                & 1  & S       & 16 (circular and hemi–sphere)\\
\cite{narang2014auditory}     & Subspace analysis    & 2014 & Humanoid           & 1  & S       & 7 (6 circular + 1 top)    \\
\cite{asano2015sound}         & Subspace analysis    & 2015 & Humanoid           & 1  & S       & 2              \\
\cite{nakamura2015interactive}& Subspace analysis    & 2015 & General            & 1  & S\,+\,M & 8 (planar)     \\
\cite{hoshiba2017design}      & Subspace analysis    & 2017 & UAV                & 1  & S       & 12 / 16         \\
\cite{nakadai2017development} & Subspace analysis    & 2017 & UAV                & 1  & S       & 12 (spherical)  \\
\cite{suzuki2017development}  & Subspace analysis    & 2017 & Mobile ground      & 1  & S       & 8 (circular)    \\
\cite{sewtz2020robust}        & Subspace analysis    & 2020 & Humanoid           & 1  & S\,+\,M & 4 (linear)      \\
\cite{manamperi2022drone}     & Subspace analysis    & 2022 & UAV                & 2  & S       & 30             \\
\cite{chen2022broadband}      & Subspace analysis    & 2022 & Mobile ground      & 3  & S       & 24 (rectangular)\\
\cite{chen2022large}          & Subspace analysis    & 2022 & Mobile ground      & 3  & S       & 16 (planar)     \\
\cite{shi2023audio}           & Subspace analysis    & 2023 & Mobile ground      & 3  & S       & 4              \\
\cite{yamada2023placement}    & Subspace analysis    & 2023 & UAV                & 2  & M       & 16 (spherical)  \\
\bottomrule
\end{tabular}}
\end{table*}

\subsection{DL-based SSL}

% The limitations of traditional SSL methods, particularly in challenging acoustic environments on one hand, and the fact that DL models could learn the acoustic features automatically, with no, or low need of preprocessing input signals\cite{grumiaux2022survey}, on the other, have motivated researchers to exploit them in SSL.

Traditional SSL methods rely on explicit mathematical models of sound propagation and array geometry. While effective in controlled conditions, they often struggle in challenging real-world environments, such as high noise levels, reverberation, multiple sound sources, and unknown or changing array configurations, as well as moving sound sources. Machine Learning (ML) and especially Deep Learning (DL) have emerged as transformative approaches in SSL by enabling models to learn intricate patterns directly from data, with no or low need of pre-processing input data \cite{grumiaux2022survey,he2021deep}, often outperforming traditional methods in challenging scenarios.
This section examines the evolution, methodologies, architectures, and innovations in deep learning-based SSL for robotics applications.

\subsubsection{Traditional Machine Learning and Neural Networks}

Before ‘‘deep’’ became fashionable, researchers already tried to map acoustic features to directions with classical machine–learning.  support-vector machines (SVMs) model is a well-known ML that has been implemented in some SSL works as well as sound event classification \cite{tran2010sound,wang2006environmental,yussif2023application}. Interestingly, in some SVM-based sound source localization the data features from traditional SSL has been used, such as SVM using TDOA features \cite{chen2011sound} or SVM based on Beamforming \cite{salvati2016weighted,salvati2016use}. $K$‐nearest neighbours (KNNs), is another classical ML model that has been used in sound source localization \cite{gadre2023comparative} and sound source classification \cite{wang2006environmental,nando2019increasing}.

Aside from traditional machine learning, which are primarily dependent on the quality and relevance of the hand-crafted features extracted from the raw audio signals (e.g., TDoA values from GCC \cite{sun2017indoor}), neural networks were able to automatically learn optimal features directly from raw multi-channel audio data or low-level spectrograms. 
At the beginning, some research used simple neural networks (NNs) in the SSL task. This could be done in the simplest form using a shallow neural network (having a single hidden layer) or NNs with multiple hidden layers, called multi-layer perceptrons (MLPs) \cite{lecun2015deep}. 
Shallow neural networks were not frequently used in SSL \cite{fu2015development}, since having only 1 layer as a hidden layer, couldn't let the model learn the sound features for localization accurately.
Song et al. \cite{song2020automatic}, used a shallow neural network with fuzzy inference, called the fuzzy neural network (FNN), for fault detection from sound signals using a plant machinery diagnosis robot (MDR).
In contrast, many SSL works used deeper neural networks, with multiple hidden layers, to effectively learn during the training phase the sound features. To the best of our knowledge, in the beginning of 2000, some early works used shallow MLPs in SSL, such as NNs with 2 hidden layers \cite{jin2000neural,pu1997neuromorphic}.
Kim et al. \cite{kim2011direction} used MLPs for both sound counting and localization, such that an MLP detected the number of sound sources, and then a different MLP localized each of the sound sources.
Also, MLP was used in two studies by Davila-Chacon et al. \cite{davila2014improving,davila2018enhanced} on a humanoid robot to estimate the azimuth angle and also to improve automatic speech recognition (ASR). 
In robotics, Youssef et al. \cite{youssef2013learning} used an MLP to estimate sound source azimuth angle by exploiting the binaural signal received in two microphones mounted on a humanoid robot. Another robotic study by Murray and Ervin \cite{murray2011neural} used an MLP for estimating elevation angle while sounds were recorded through artificial pinnae, mimicking the human auditory system.
In another robotic study by He et al. \cite{he2018deep}, they proposed an MLP-based model for simultaneously detecting and localizing multiple sound sources in human-robot interaction scenarios, and they showed their model outperformed traditional SSL, such as MUSIC-based. Similarly, Takeda and Komatani used MLP in their works and showed MLP model outperforms MUSIC in SSL in studies with a robotic theme \cite{takeda2016discriminative,takeda2016sound,takeda2017unsupervised}.

It is important to note that the neural network architecture in each study is different from each other and has many hyperparameters. For example, the activation function in each layer, the number of hidden layers and nodes in each layer, as well as the type of input data, i.e. could be raw signal, spectrogram, or sound features captured from traditional SSL, could be different for each NN. Therefore, hyperparameter selection through an ablation study could help in achieving the best architecture. Also, another challenge in ML-based SSL is to ensure the quantity and quality of data is sufficient in training, such that the model could learn effectively in the training phase and avoid some common issues, such as overfitting \cite{bianco2019machine}.

\subsubsection{Convolutional Neural Networks (CNNs)}

% Localization performance jumped when researchers fed networks \emph{structured} inputs—STFT magnitude/phase stacks, cyclic GCC images, or 3-D voxelised FOA (first-order Ambisonics) tensors—and used convolutional neural networks (CNNs) to learn spatial kernels.  Hirvonen’s 2015 work on iCub marked the turning point: a simple 2-D CNN operating on left–right spectrogram pairs halved the binaural error compared to hand-coded ILD + ITD rules.  Subsequent designs exploited 3-D convolutions to fuse time–frequency–channel dimensions, enabling eight-mic service robots such as SoftBank’s NAO or Mitsubishi’s HSR to achieve below-$2^{\circ}$ mean error in cluttered apartments.

% While early machine learning models might have treated sound data as simple numerical vectors, a pivotal shift in SSL came with the adoption of Convolutional Neural Networks (CNNs). CNNs are popular class of DNNs widely used for pattern recognition due to their property of being translation equivariant

A convolutional neural network is a famous DL architecture composed of learnable convolution kernels, interleaved with non–linear activations and, typically, pooling or normalization layers \cite{lecun2015deep}. The defining feature is weight sharing: the same kernel is slid across the input, allowing the network to detect local patterns regardless of their absolute position. CNNs were first popularised for image
analysis \cite{krizhevsky2012imagenet}, yet their inductive biases, translation invariance, and local receptive fields map naturally to spectro-temporal audio representations.
% When an STFT, Mel-spectrogram, or a stack of GCC-PHAT frames is laid out as a 2- or 3-D tensor (\textit{frequency $\times$ time $\times$ channel}), convolutional filters become trainable beamformers that learn to highlight inter-channel phase constellations and amplitude modulations characteristic of specific directions of arrival.
Compared to analytic TDOA or MUSIC pipelines, CNNs require no hand-crafted feature selection \cite{desai2022review}, can fuse information across multiple microphones and hundreds of frequency bins in a single forward pass, and can be trained to ignore ego-noise unique to a given platform. Their computation is a chain of dense tensor operations that map efficiently onto embedded GPUs, making real-time deployment feasible even on small service robots.

As pointed out in the survey by Grumiaux et al. \cite{grumiaux2022survey}, in 2015, the first CNN-based SSL was done by Hirvonen \cite{hirvonen2015classification}. In this study a CNN model was trained to classify an audio signal containing one speech or musical source into spatial regions as a classification task. Some studies used raw audio waveforms as input of the CNN architecture \cite{vera2018towards,suvorov2018deep,huang2021sseldnet}. This approach eliminates the need for hand-crafted features but may require larger models and more training data. However, other CNN-based SSL studies typically used the input in the form of a multichannel short-time Fourier transform (STFT) spectrogram \cite{vincent2018audio,chakrabarty2019multi}. These multichannel spectrograms are typically used as 3D tensors, with one dimension for time (or frames), one for frequency (bins), and one for channel.
Some works, used the feature of conventional SSL works, e.g. 2D images of DoA feature extracted by beamforming \cite{wu2021sound} or GCC features \cite{ferguson2018sound,butt2022active}, as an input of CNN architecture .
Regarding the convolutional layers as an important hidden layer in CNNs, Conv1D and Conv2D, as well as 3D convolutional have been used in SSL works. It is generally conceived that the higher dimension can lead to better capture of the feature while it adds computational cost \cite{krause2021comparison}. Even though, some works suggest having a hybrid of convolutional dimensions in SSL \cite{diaz2020robust}, e.g. a set of Conv2D layers for frame-wise feature extraction followed by several Conv1D layers in the time dimension for temporal aggregation \cite{bologni2021acoustic}.

In robotics, several pioneering studies demonstrate the effectiveness of CNNs in SSL tasks, particularly in overcoming the limitations of traditional methods in complex, real-world scenarios. Notably, these approaches leverage CNNs to map binaural or multi-microphone audio features directly to sound source locations or directions.
Nguyen et al. \cite{nguyen2018autonomous} explored an autonomous sensorimotor learning framework for a humanoid robot (iCub) to localize speech using a binaural hearing system. Their contribution involved an automated data collection and labeling procedure, and they successfully trained a CNN with white noise to map audio features to the relative angle of a sound source. Crucially, their experiments with speech signals showed the CNN's capability to localize sources even without explicit spectral feature handling, highlighting the network's inherent ability to learn robust mappings. 
Similarly, \citet{boztas2023sound} focused on providing auditory perception for humanoid robots, emphasizing the importance of localizing moving sound sources in scenarios where cameras might be ineffective. By recording audio from four microphones on a robot's head and combining it with odometry data, the study demonstrated that a CNN could accurately estimate the location of a moving sound source, thus validating the robot's ability to sense sound positions with high accuracy, akin to living creatures.
Furthermore, CNN-based SSL models are pushing the boundaries to address more complex robotic challenges.  \citet{he2018deep} proposed using neural networks for the simultaneous detection and localization of multiple sound sources in human-robot interaction, moving beyond the single-source focus of many conventional methods. They introduced a likelihood-based output encoding for arbitrary numbers of sources and investigated sub-band cross-correlation features, demonstrating that their proposed methods significantly outperform traditional spatial spectrum-based approaches on real robot data. 
\citet{jo2025sound} presented a three-step deep learning-based SSL method tailored for human-robot interaction (HRI) in intelligent robot environments. Their approach prioritizes minimizing noise and reverberation by extracting Excitation Source Information (ESI) using linear prediction residual to focus on the original sound components. Subsequently, GCC-PHAT is applied to cross-correlation signals from adjacent microphone pairs, with these processed single-channel, multi-input cross-correlation signals then fed into a CNN. This design, which avoids complex multi-channel inputs, allows the CNN architecture to independently learn TDOA information and classify the sound source's location.
While not directly tested on a robot, \citet{pang2019multitask} introduced a multitask learning approach using a Time-Frequency CNN (TF-CNN) for binaural SSL, aiming to simultaneously localize both azimuth and elevation under unknown acoustic conditions. Their method extracts interaural phase and level differences as input, mapping them to sound direction, and shows promising localization performance, explicitly stating its usefulness for human-robot interaction. Similarly, \citet{ko2022real} proposed a multi-stream CNN model for real-time SSL on low-power IoT devices, including its application to camera-based humanoid robots. Their model processes multi-channel acoustic data through parallel CNN layers to capture frequency-specific delay patterns, achieving high accuracy on noisy data with low processing time, further emphasizing its applicability for robots mimicking human reactions in crowded environments.
CNN-based SSL has also been tested in a multi-robot configuration. In this regard, \citet{mjaid2023ai} introduced AudioLocNet, a novel lightweight CNN that frames SSL as a classification task over a polar grid, letting each robot infer both azimuth and range to locate and communicate with each other.

\subsubsection{Convolutional Recurrent Neural Networks (CRNNs)}

A CNN treats each input window independently; it excels at learning spatial-spectral patterns, but it has no mechanism to remember how those patterns evolve over time.  Sound localization in the wild, however, is inherently temporal.  Robots rotate their heads, sources start and stop speaking, and early reflections arrive a few milliseconds after the direct path. 
To deal with temporal dependency, recurrent neural network (RNN) architectures are designed by adding state vectors that are updated at every time step, enabling the model to maintain a memory of past observations and to capture temporal dependencies that a vanilla CNN cannot represent.  In the context of sound localization an RNN ingests a sequence of time–frequency frames, propagates information through gated recurrent units (GRUs) \cite{cho2014learning} or long short-term memory (LSTM) cells \cite{hochreiter1997long}, and outputs either frame-wise direction estimates or a smoothed trajectory of source positions.  Regarding RNNs in SSL, few works such as work by Nguyen et al.,\cite{nguyen2021general} and Wang et al. \cite{wang2018robust} were used RNNs without combining it with CNNs \cite{grumiaux2022survey}. The former work implemented an RNN to match and fuse sound event detection (SED) and DOA predictions, and the latter one used a bidirectional LSTM to identify speech dominant time-frequency units for DOA estimation.
% demonstrated that a two-layer bidirectional LSTM trained on binaural features could track a moving talker mounted on a TurtleBot with lower latency than a particle filter, while Chakrabarty and Habets showed that a single-source DoA regression network based on stacked GRUs retained accuracy in rooms with \(T_{60}\) up to 0.7\,s where GCC-PHAT degraded substantially\cite{chakrabarty2017rnn}.
Despite these successes, pure RNN solutions remained limited in SSL (e.g., two SSL works in robotics \cite{jin2020rnn,andra2020portable}) because the recurrent layers themselves could not learn spatial filters or phase relations from raw spectra.  This bottleneck set the stage for convolutional–recurrent hybrids (CRNNs), which delegate spatial feature extraction to a front-end CNN and let the RNN specialise in temporal reasoning, thus combining the best of both worlds.

CRNNs have been regularly implemented for SSL since 2018 \cite{grumiaux2022survey}. Having some convolutional layers following a recurrent unit, such as a bidirectional gated recurrent unit (BGRU) or LSTM, and a fully connected layer is a typical CRNN architecture in SSL. This kind of model have been showed to be robustly able to detect overlapped sounds and a noisy environment \cite{adavanne2018sound,lu2019sound}.
Regarding the localization of multiple sound sources, some CRNN-based SSL works used a separate CRNN for sound source counting as well as one for localization \cite{tian2020multiple}, while some assumed the prior knowledge of sound sources and localize each with a trained model \cite{jalayer2024convlstm}, and some simultaneously localize multiple sound sources within a single CRNN model \cite{perotin2019crnn,grumiaux2021improved}.

In robotics, CRNNs have been recently (after 2020) used in SSL tasks. For example, \citet{kim2021mimo} addressed the critical challenge of ego noise generated by mobile robots (robot vacuum cleaner), which significantly degrades SSL performance. Their approach proposed a multi-input multi-output (MIMO) noise suppression algorithm built upon an extended time-domain audio separation network (TasNet \cite{han2020real}) architecture. They evaluated their work by the CNN (U-Net \cite{ronneberger2015u}) and CRNN (by adding an LSTM block) models across different noise levels. Aside from Kim et al. works, other CRNN architecture in SSL have been used indirectly in robotic application. For instance, Mack et al. \cite{mack2020signal} implemented a Conv-BiLSTM model to estimate the DOA angles based on spectrogram inputs of human voice and artificial noise with possible applications in robotics. Jalayer et al. \cite{jalayer2024convlstm} implemented CovLSTM-based SSL for localization of humans and machines (CNC machines) using raw audio signals in an industrial noisy environment that can be applied for scenarios with industrial robots in the scene. Two studies by Altayeva et al. \cite{altayeva2023convolutional} and Akter et al. \cite{akter2025hybrid} implemented ConvLSTM-based sound detection and classification for various sound events in an urban environment, which could be applied in security robots in urban areas. Varnita et al. \cite{varnita2024precision} similarly implemented ConvLSTM-based sound event localization and detection (SELD), which could be practically used for accurate sound-based navigation and awareness of robots regarding the events.

\subsubsection{Attention-based SSL}

The intuition behind {\it attention} is simple yet powerful: rather than processing an input sequence with a fixed, locality–biased kernel, a model can learn to {\it select} and {\it weight} the time–frequency regions that matter most for the task at hand.  Introduced by Bahdanau et al. \cite{bahdanau2014neural} for machine translation and generalized in the Transformer of Vaswani et al. \cite{vaswani2017attention}, attention replaces recurrence with a data–driven affinity matrix that relates every element of the input to every other element.  Each output representation is thus a context–aware combination of the entire input, enabling the network to capture long-range dependencies and intricate phase relationships across microphone channels that are essential for precise sound localization \cite{grumiaux2022survey}.

The first wave of attention-augmented SSL networks (after 2020) kept the overall CRNN structure and simply inserted a self-attention layer after the recurrent stack.  Phan et al. \cite{phan2020multitask} and Schymura et al.\cite{schymura2021exploiting} reported consistent gains over the CRNN baseline: localization error decreased while the models learned to focus automatically on time frames where sources were active. Mack et al.\cite{mack2020signal} placed two successive attention blocks, the first operating directly on phase spectrograms, the second on convolutional feature maps, to generate dynamic masks that emphasise frequency bins dominated by the target speaker.
Following the broader “{\it attention-is-all-you-need}” trend, fully Transformer-style encoders are now replacing recurrent back-ends altogether.  Multi-head self-attention (MHSA) layers were coupled with lightweight CNN front ends by Emmanuel et al. \cite{emmanuel2021multi} and Yalta et al. \cite{yalta2021hitachi}, cutting computation time while winning the SELD track. Many SSL research in recent years directly implemented Transformers in SSL \cite{zhang2024dynamic,zhang2024novel,chen2025hybrid}, for instnae the most recent one by  Zhang et al. \cite{zhang2025multiple} extend this trend by coupling a CNN front-end with a Transformer encoder that attends over sub-band spatial-spectrum features, enabling the network to identify true DOA peaks and suppress spurious ones, and thereby delivering state-of-the-art localization accuracy for multiple simultaneous sources under heavy noise and reverberation.

Although Transformer-style architectures already set the performance bar on SSL benchmarks, documented examples of their on-board use in mobile or humanoid robots remain scarce. Yet the very property that makes attention models excel in controlled evaluations, namely, their ability to focus dynamically on the most informative time-frequency cues, addresses exactly the challenges faced by robots in the wild. In a bustling factory or a crowded café, an attentive SSL front-end can learn to highlight the spectral traces of a target speaker while suppressing background chatter, motor ego-noise, and sporadic impacts, thus preserving a stable acoustic focus for downstream speech recognition, navigation, or event detection. Because multi-head self-attention offers a global receptive field and permutation-invariant decoding, it naturally accommodates multi-source tracking without handcrafted heuristics, making it a promising cornerstone for robots that must localize critical events in real time and interact seamlessly with humans in complex soundscapes.
% While the computational load of full Transformers is still higher than that of pure CNNs, several pragmatic strategies already make attention-based SSL feasible on embedded robotic hardware.
% For instance, sparse attention variants reduce the quadratic complexity of the self-attention matrix \cite{lin2022survey}, cutting both latency and RAM requirements without a measurable loss in localization accuracy \cite{mu2023improving}. Also, weight-sharing across heads and post-training quantization \cite{liu2021post} enable multi-head self-attention to execute comfortably within the power envelope of Jetson-class GPUs or Edge-TPU accelerators carried by many service and inspection robots. Given their robustness to overlapping sources, ability to handle variable array geometries, and synergy with other sensory modalities, attention-based networks are poised to become a core building block of next-generation robotic audition.
Admittedly, full Transformers demand more compute and memory than lightweight CNNs, but recent engineering advances bring their inference cost within reach of embedded platforms. Sparse attention variants reduce the quadratic complexity of self-attention, slashing both latency and RAM with negligible loss in angular accuracy \cite{lin2022survey,mu2023improving}. Further savings arise from weight-sharing across heads and post-training quantization \cite{liu2021post}, allowing multi-head self-attention to execute comfortably on Jetson-class GPUs or Edge-TPUs already found on many service, inspection, and aerial robots. These techniques suggest that attention-driven SSL is poised to become a practical and powerful component of next-generation robotic audition systems.

% To better provide the information regarding papers using DL and ML-based SSL in robotics, we listed them in Table \ref{tab:ml_ssl_robots}. As can be seen using traditional DL and ML (e.g., MLP) were more common at the beginning of the last decade, while lately papers have used more CNNs, RNNs, and CRNNs for SSL in robotics. Interestingly, the transformers yet to be implemented in SSL in robotics due to their performance superiority over most architectures in SSL.
% Since CNNs, RNNs, and CRNNs architectures allows capturing spatial-spectral patterns and temporal features in acoustic inputs, using them let researchers explore SSL in the presence of multiple sound sources (e.g. \cite{jalayer2024convlstm}, and also moving sound sources \cite{jin2020rnn,boztas2023sound,jalayer2024convlstm}.
Table \ref{tab:ml_ssl_robots} summarizes the studies that apply machine- and deep-learning techniques to SSL on robotic platforms. Early work (2013–2017) relied mainly on shallow neural networks and MLPs or on classical feature–reduction methods and machine learning (e.g., PCA and SVM). From 2017 onward, however, the field has shifted decisively toward convolutional, recurrent, and hybrid (CRNN) architectures. These models can jointly exploit the spatial–spectral structure of microphone signals and their temporal dynamics, enabling, for example, localization of several simultaneous active sound sources \cite{jalayer2024convlstm} or tracking of moving emitters \cite{jin2020rnn,boztas2023sound,jalayer2024convlstm}.
Notably, to the best of our knowledge, Transformer-based networks—now state-of-the-art in many audio tasks—have not yet been adopted for robotic SSL, leaving a promising avenue for future research.

\begin{table*}[h]
    \scriptsize
    \centering
    \begin{threeparttable}
    \caption{Machine- and Deep-Learning–based SSL studies on robotic platforms}
    \label{tab:ml_ssl_robots}
    \begin{tabular}{@{}l c c l c c c@{}}
        \toprule
        \textbf{Paper} & \textbf{ML/DL Method} & \textbf{Year} & \textbf{Robot Type} &
        \textbf{Max.~active sources} & \textbf{S / M} & \textbf{No. of Microphones (geometry)} \\ \midrule
        \cite{youssef2013learning}      & MLP               & 2013 & Humanoid          & 1 & S           & 2 (binaural) \\
        \cite{keyrouz2014advanced}      & PCA{+}DFE         & 2014 & Humanoid          & 1 & S           & 2 (binaural) \\
        \cite{davila2014improving}      & MLP               & 2014 & Humanoid          & 1 & S           & 2 (binaural) \\
        \cite{deleforge2015acoustic}    & PPAM              & 2015 & Humanoid          & 3 & S           & 2 (binaural) \\
        \cite{takeda2016discriminative} & MLP               & 2016 & General           & 2 & S           & 2 (binaural) \\
        \cite{takeda2016sound}          & MLP               & 2016 & General           & 1 & S           & 2 (binaural) \\
        \cite{sun2017indoor}            & SVM               & 2017 & General           & 1 & S           & 6            \\
        \cite{takeda2017unsupervised}   & MLP               & 2017 & Humanoid          & 1 & S           & 2 (binaural) \\
        \cite{he2018deep}               & CNN               & 2018 & Humanoid          & 2 & S           & 4            \\
        \cite{davila2018enhanced}       & MLP               & 2018 & Humanoid          & 2 & S           & 2 (binaural) \\
        \cite{pang2019multitask}        & CNN               & 2019 & General           & 1 & S           & 2 (binaural) \\
        \cite{song2020automatic}        & FNN               & 2020 & Mobile ground     & 1 & S           & 8            \\
        \cite{andra2020portable}        & LSTM              & 2020 & Mobile ground     & 1 & S           & 6            \\
        \cite{jin2020rnn}               & RNN               & 2020 & Mobile ground     & 1 & M           & 20           \\
        \cite{gonzalez2021self}         & CNN               & 2021 & Humanoid          & 1 & S           & 2 (binaural) \\
        \cite{kim2021mimo}              & CRNN              & 2021 & Mobile ground     & 1 & S           & 4 (tetrahedral) \\
        \cite{butt2022active}           & CNN               & 2022 & Humanoid          & 1 & S           & 2 (binaural) \\
        \cite{boztas2023sound}          & MLP/CNN/LSTM      & 2023 & Humanoid          & 1 & M           & 4            \\
        \cite{jalayer2024convlstm}      & CNN + LSTM              & 2024 & General           & 5 & S{+}M       & 3            \\
        \cite{jo2025sound}              & CNN               & 2025 & Mobile ground     & 1 & S           & 3            \\ \bottomrule
    \end{tabular}
    \begin{tablenotes}[flushleft]
    \footnotesize
    \item \textbf{Acronyms:} PCA – Principal Component Analysis; DFE – Diffused Field Equalization; PPAM – Probabilistic Piecewise Affine Mapping.
    \end{tablenotes}
    \end{threeparttable}
\end{table*}

% Principal Component Analysis (PCA), Diffused Field Equalization (DFE), Probabilistic Piecewise Affine Mapping (PPAM),

\section{Data and Learning strategies}
\label{sec:data+learning}
% The efficacy of modern Sound Source Localization (SSL) systems, particularly those employing Deep Learning (DL) models, hinges critically on robust data and effective learning strategies. For robotic applications, these aspects are especially nuanced due to the inherent complexities of real-world operational environments.

Deep-learning sound-source localization requires extensive and diverse corpora as well as training paradigms that bridge the gap between laboratory conditions and everyday robotics. Mobile robots face motor ego-noise, rapidly changing geometries, overlapping talkers, and strong reverberation. We review (i) how training data are generated or collected, (ii) the augmentation schemes that mitigate over-fitting, and (iii) supervised, semi-/weakly-supervised, self-supervised, and transfer-learning paradigms that turn those data into robust models.

\subsection{Data}

\textbf{Simulation pipelines:} Most systems bootstrap with synthetic data because collecting ground-truth directions for every time frame is costly. The standard recipe convolves dry audio (a reverberation-free data) with room-impulse responses (RIRs) generated by an image-source method (ISM) \cite{allen1979image} or its GPU-accelerated variants \cite{diaz2021gpurir}.
Open-source libraries such as Habets’ RIR generator \cite{habets2006room}, Pyroomacoustics \cite{scheibler2018pyroomacoustics}, and ROOMSIM \cite{campbell2005matlab,schimmel2009fast} can render millions of multichannel RIRs spanning room sizes, reverberation times and source–array poses \cite{grumiaux2022survey}. Models trained on such corpora generalize surprisingly well \cite{grumiaux2021improved,varanasi2020deep}, e.g., the VAST dataset \cite{gaultier2017vast} covers a vast variety of different geometries simulated in ROOMSIM and showed virtually-learned mappings on this dataset generalize to real test data. Although they have some limitations, for example, a simulated acoustic room (e.g., generated by Pyroomacoustics) is inherently unable to generate external diffuse noise and simulate obstacles or separating walls within the simulated room \cite{cheng2020mass,jalayer2024convlstm}.
Dry-signal choice matters: mixing speech, noise, and sound events outperforms noise-only training (Krause et al. \cite{krause2021data}). For robotics, to narrow the gap between simulation to reality, specific considerations such as ego-noise (generated by the robot itself), simulating recording from moving microphones \cite{evers2016localization}, in case microphones are mounted on a mobile robot, should be taken into account. Achieving a high-fidelity simulation of SSL in robotics can avoid risky and costly field trials.

\textbf{Recorded datasets:} Real corpora remain indispensable for evaluation. 
Regarding the data collection the emergence of some worldwide challenges organized in the past years has motivated publicly sharing the datasets.
One of the most widely used real‐recorded benchmarks for modern localization networks are the sound-event localization and detection (SELD) datasets released by the \emph{Detection and Classification of Acoustic Scenes and Events} (DCASE) challenge \cite{mesaros2025decade}.  Open science is a guiding principle of DCASE: every task ships an openly licensed dataset, a baseline system, and a fixed evaluation protocol, enabling reproducible comparison across systems and years. For SSL, the relevant tasks included the localization of static sound sources \cite{adavanne2019multi} as well as moving ones \cite{politis2020overview,politis2021dataset,politis2022starss22}. The latest development of the SELD task include using audio-visual input \cite{shimada2023starss23} and, in 2024, distance estimation \cite{krause2024sound}.
These datasets were recorded in the real-world reverberant and noisy environments containing different sound types (e.g., human speech and barking dog sounds). As highlighted in the survey by Grumiaux et al. \cite{grumiaux2022survey}, the datasets of the DCASE challenges have become the benchmark for deep learning-based SSL, e.g., \cite{yeow2024squeeze,shimada2020sound}.  Although the recordings are not captured on mobile robots, the datasets’ noisy, reverberant, multi-source scenarios mirror many robotic deployments and therefore could serve as an invaluable data source for some robotic research.
The acoustic source Localization and Tracking (LOCATA) challenge \cite{evers2020locata} provides another comprehensive data corpus encompassing scenarios from a single static source to multiple moving speakers, using various microphone arrays (from a 15-microphone planar array to a 12-microphone robot head array and even binaural hearing aids). This dataset could be very beneficial for robotic SSL since it includes the data captured from A pseudo-spherical array with 12 microphones integrated into a prototype head for the humanoid robot NAO.
% TAU Spatial Sound Event sets from DCASE provide strongly-labelled FOA mixtures in real homes and offices \cite{politis2021dcase}. 

Aside from those well-established datasets in acoustic challenges, some robotics-specific datasets have also been published.
One of the examples is the SSLR data introduced by He et al.,\cite{he2018deep}. In this dataset, all recordings were made with the four head-mounted microphones of a Pepper humanoid, so every recording includes the robot’s own fan noise. Two subsets are available: a loudspeaker subset in which Pepper automatically pans its head while speech is played from random positions, and a human-interaction subset that captures single and overlapping utterances during live dialogues.
Both subsets provide motion-capture ground truth and voice-activity labels, making SSLR particularly suitable for human–robot-interaction studies.
Another dataset by \cite{li2016reverberant}, designed for robot audition, was recorded across four real-world environments with varying reverberation times, utilizing a NAO humanoid robot equipped with microphones on its head. It comprises Audio-only data, i.e., speech emitted via loudspeaker in the lab, fixed distances, 360° azimuth/elevation range, and Audio-visual data, i.e., speech within the robot's camera field-of-view, both predominantly affected by the robot's self-generated fan noise.
For aerial robotics the most complete resource is DREGON, published by Strauss et al.,\cite{strauss2018dregon}.
Here an eight-microphone cube is suspended beneath a quadrotor; flights are tracked by a Vicon system and accompanied by synchronized IMU and motor-speed logs.
The corpus combines challenging in-flight speech or noise at negative SNRs with separate ego-noise flights and semi-anechoic “clean” loud-speaker recordings, so researchers can test localization, noise suppression, and data synthesis with the same material.
Wang et al.,\cite{wang2019audio} extended the aerial scenario with an audio-visual quadcopter (AVQ) dataset.
The 50-minute collection is divided into two subsets. In the first, up to two talkers stand at nine predefined locations between two and six meters from the drone while speech and rotor noise are recorded separately, allowing controlled SNR studies.
In the second subset a loud-speaker is carried along three-minute trajectories, giving moving-source material at varying thrust levels.
Video frames at 30 fps provide the ground-truth angles for both subsets.
Most recently Jekaterynczuk et al.\cite{jekaterynczuk2025uavirbase} presented UaVirBASE, a publicly downloadable database aimed at ground-based drone monitoring.
It contains high-quality recordings of a single UAV captured from many distances, heights, and azimuths, and includes detailed metadata such as array coordinates and environmental conditions.
A baseline mel-spectrogram DNN trained on UaVirBASE achieves roughly half-meter mean error in range and about one degree in azimuth, demonstrating the dataset’s suitability for acoustic surveillance applications.
Taken together, SSLR, DREGON, AVQ, and UaVirBASE provide complementary test beds that include ego-noise, platform motion, and realistic microphone layouts, conditions that conventional laboratory corpora and purely simulated data rarely reproduce in full.

\textbf{Data augmentation.}
The performance of deep learning-based SSL models, especially in complex robotic applications, is highly dependent on the quantity and diversity of training data. To overcome limitations in real-world data availability and mitigate the potential for models to overfit to simulated conditions, data augmentation techniques are indispensable. These methods create new training examples from existing data, enriching the dataset without requiring additional physical recordings, and ultimately leading to more robust and generalizable SSL models for robots.
Various data augmentation strategies have been explored, many inspired by their success in other domains. For instance, techniques like SpecAugment \cite{zhang2019data}, originally used for speech recognition \cite{park2019specaugment}, involve masking certain time frames or frequency bands within spectrograms. This teaches the model to focus on more fundamental spatial cues rather than relying on specific spectral content, a valuable trait for robots encountering diverse sound characteristics. Other methods leverage the properties of multi-channel audio formats, such as First Order Ambisonics (FOA). Mazzon et al. \cite{mazzon2019first} proposed techniques like channel swapping or inversion and label-oriented rotation, which effectively generate new sound source locations and orientations by transforming the FOA channels and their corresponding labels. This allows a robot to learn how sound propagates and presents itself from a wider range of directions than might be practically recordable. More broadly, methods like Mixup \cite{pratik2019sound} generate new training samples by creating convex combinations of existing data pairs, while techniques like pitch shifting \cite{noh2019three} and random mixing of multiple training signals \cite{wang2023four,takahashi2016deep} create novel mixtures, simulating scenarios with varying speaker characteristics or multiple overlapping sound sources.
For robotics, these data augmentation techniques translate directly into significant practical advantages. By exposing SSL models to a vastly expanded and diversified set of synthetic acoustic environments, robots can develop robust "acoustic perception" that is less susceptible to the unpredictability of real-world deployment. Robots equipped with models trained on augmented data will be better at localizing sources even in highly reverberant rooms, amidst unforeseen background noises, or when facing multiple simultaneous speakers.

\subsection{Learning paradigms}

The effectiveness of deep neural networks in Sound Source Localization heavily depends on the chosen learning strategy, which dictates how the model acquires knowledge from data. While most SSL systems rely on supervised learning \cite{grumiaux2022survey}, the limitations of labeled data in complex robot environments necessitate exploring semi-supervised, weakly supervised, and transfer learning approaches.

\textbf{Supervised Learning:} Supervised learning is the predominant paradigm in DL-based SSL. It involves training a neural network on a dataset where each input (e.g., multi-channel audio features) is paired with a corresponding ground truth output (the "label"), the known position or direction of the sound source. The network learns by minimizing a cost function (or loss function), which quantifies the discrepancy between its predicted output and the true label.

For robotic SSL, supervised learning typically manifests in two primary ways:
\begin{itemize}
    \item Classification: When the sound source's location is discretized into angular bins (e.g., 5-degree sectors around the robot). Here, a softmax activation function is often used at the output, and the model minimizes the categorical cross-entropy loss. This is suitable for tasks where a robot needs to identify which general direction a sound is coming from \cite{jo2025sound}.
    \item Regression: When the goal is to predict continuous values for the sound source's azimuth, elevation, or 3D Cartesian coordinates. In this case, the mean square error (MSE) is the most common choice for the cost function\cite{he2021neural}. This enables robots to pinpoint locations with greater precision. While MSE is prevalent, other metrics like angular error or L1-norm are sometimes employed to capture specific aspects of localization accuracy \cite{jenrungrot2020cone}.
\end{itemize}

The primary limitation of supervised learning for robotic applications is its insatiable demand for large amounts of accurately labeled training data. Collecting such data in real-world robotic environments is incredibly resource-intensive, as it requires meticulously tracking sound sources and robot poses. Existing real-world SSL datasets for robotics are often limited in size and variety, making them insufficient for robust DL model training. To address this, supervised learning is often augmented with extensive data simulation and data augmentation techniques, which synthetically expand the dataset's diversity.

\textbf{Beyond Full Supervision:} To cope with the scarcity of labeled real-world data and enhance model robustness to unseen conditions, researchers leverage strategies that go beyond purely supervised learning:

\begin{itemize}
    \item Semi-Supervised Learning: This approach combines both labeled and unlabeled data for training \cite{grumiaux2022survey}. The core idea is to perform part of the learning in a supervised manner (using available labels) and another part in an unsupervised manner (learning from unlabeled data) \cite{hu2020semi}. For robotics, this is highly valuable because robots can continuously collect vast amounts of unlabeled audio data during their operation. Semi-supervised learning methods can fine-tune a network pre-trained on labeled data (often simulated data), adapting it to real-world conditions without requiring exhaustive manual labeling. Techniques such as minimizing overall entropy (e.g., SSL work by Takeda et al. \cite{takeda2018unsupervised}) or employing generative models (e.g., SSL work by Bianco et al. \cite{bianco2020semi}) that learn underlying data distributions have been proposed. Adversarial training (e.g. SSL work by Le Moing et al. \cite{le2021data}) is another example, where a discriminator network tries to distinguish real from simulated data, while the SSL network learns to "fool" it by producing realistic outputs from simulated inputs, thus adapting to real-world acoustic characteristics. This allows a robot to refine its SSL capabilities based on its own experiences in the operational environment, even if precise ground truth is unavailable.
    \item Weakly Supervised Learning: In this paradigm, the training data comes with "weak" or imprecise labels, rather than detailed ground truth in many domains when labeling is costly or challenging \cite{ren2023weakly}. For SSL, this might mean only knowing the number of sound sources present, or having a rough idea of their general location, without exact coordinates. Models are designed with specialized cost functions that can account for these less precise labels. For instance, He et al. \cite{he2019adaptation} fine-tuned networks using weak labels representing the known number of sources, which helped regularize predictions. Other approaches, like those using triplet loss functions by Opochinsky et al. \cite{opochinsky2019deep}, involve training the network to correctly infer the relative positions of sound sources (e.g., that one source is closer than another), even if their absolute coordinates aren't provided. For robotics, weakly supervised learning offers a pragmatic solution when collecting precise ground truth labels is impractical, allowing robots to learn from more easily obtainable, albeit less granular, supervisory signals. This means a robot could learn to localize effectively just by being told, for example, "there's a sound coming from that general direction" rather than requiring exact coordinates, making deployment and ongoing learning more feasible.
\end{itemize}

\textbf{Transfer Learning:} Transfer learning is a pivotal strategy that addresses the common challenge of data scarcity in specific target domains \cite{niu2021decade}, a particularly pertinent issue for robotic SSL. Rather than training a deep neural network from scratch on a limited robot-specific dataset, transfer learning involves leveraging knowledge acquired by a model pre-trained on a related, typically larger, source dataset. This strategy offers significant advantages for robotic applications.

The typical workflow for transfer learning in SSL involves two main phases. First, a base model is pre-trained on a vast dataset, which consists of general acoustic scenes. In this regard, some works, such as SSL studies by Nguyen et al. and Park et al. \cite{nguyen2021general,park2021many}, pre-trained their model on a large realistic dataset (e.g. audio datasets\citep{gemmeke2017audio}), and some (e.g., Zhang et al.\cite{zhang2025sound}) use simulation to generate the training data for pretraining. This initial training allows the network to learn rich, generalized representations of sound features and spatial cues. Second, the pre-trained model is then adapted to the specific SSL task through fine-tuning. This fine-tuning process involves further training the model on a comparatively smaller dataset, such as a robot-specific dataset, often with a reduced learning rate to preserve the learned general knowledge. During fine-tuning, selected layers of the pre-trained network might be frozen (acting as fixed feature extractors), or the entire network might be updated, allowing it to specialize in the specific domain, for example, the robot's acoustic environment, its ego-noise characteristics, and the unique geometry of its microphone array.

For robotics, transfer learning is highly advantageous. It significantly reduces the amount of labeled robot-specific data required, thereby accelerating development and deployment cycles. A robot can benefit from models pre-trained on generic acoustic datasets or other robotic datasets, such as SSLR, DREGON, AVQ, and UaVirBASE datasets which are detailed before, then quickly adapt to its unique operational environment with minimal new data. This strategy also improves model generalization capabilities, as the network benefits from the broader knowledge base of the pre-training data, making it more robust to unforeseen acoustic conditions, various sound sources, and environmental variations that a robot might encounter in real-world scenarios. Consequently, transfer learning stands as a powerful enabler for rapidly deploying accurate and robust SSL capabilities on diverse robotic platforms.

\section{Applications of SSL in robotics}
\label{sec:apps}

Sound Source Localization (SSL) is a fundamental capability that significantly enhances the autonomy, perceptual awareness, and interactive abilities of robotic systems. Its applications span a wide array of robot types and operational domains, enabling robust and versatile perception in complex, real-world environments. To better illustrate where SSL has already proven its worth, we categorize the papers reviewed in this study first by the robotic platform employed, namely humanoid robots, mobile ground robotic platforms, and UAVs. Subsequently, we delineate their applications based on motivating domains.

\subsection{Categorization by Robot Type}

\begin{figure*}[htbp]
    \centering
    \includegraphics[width=0.7\textwidth]{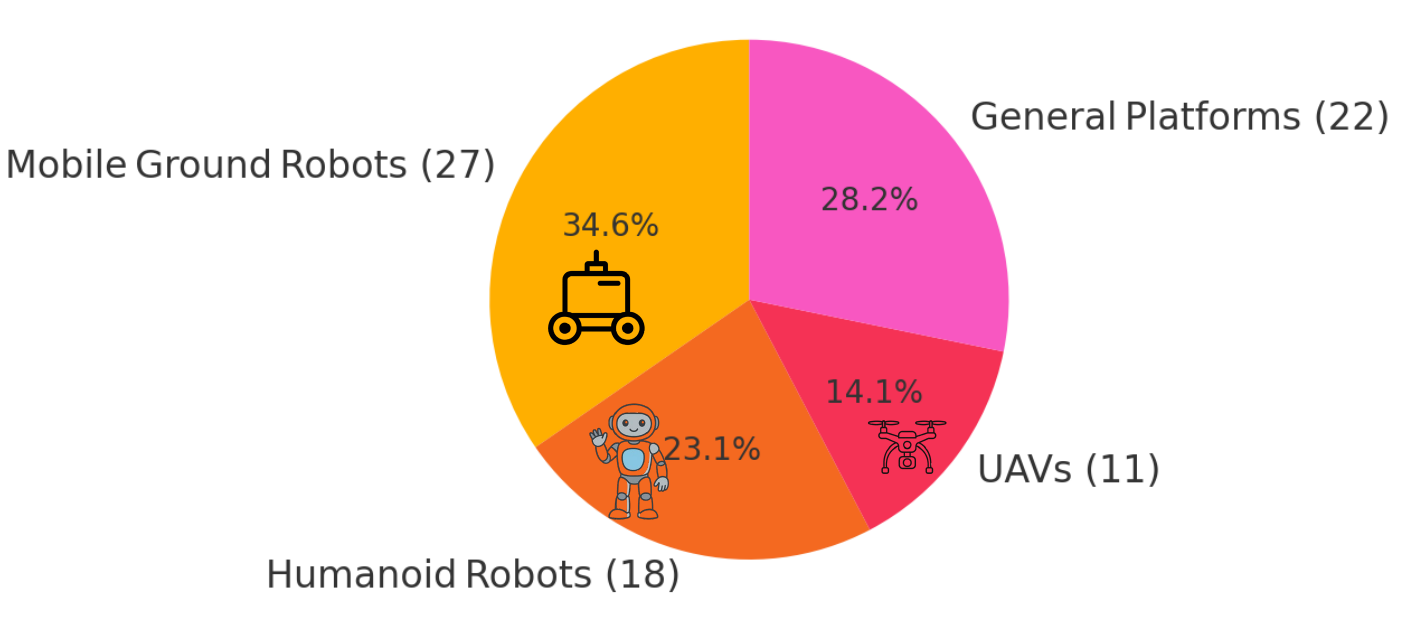} % Or .pdf, .jpg etc.
    \caption{Distribution of reviewed SSL studies by robot types.}
    \label{fig:robot_type_pie_chart} % This is the label you'd use in \ref{}
\end{figure*}

The integration of SSL capabilities varies significantly depending on the physical characteristics and primary functions of different robotic platforms. From the 78 robotic SSL papers analyzed, the predominant robot types fall into three categories: mobile ground robotic platforms, humanoid robots, and UAVs. Additionally, a notable portion of studies proposed SSL methods generalizable across platforms without specifying a particular robot type for implementation, which we categorize as "General Robotic Platforms". The distribution of these categories among the reviewed papers is depicted in Figure \ref{fig:robot_type_pie_chart}.

\begin{figure*}
    
    \centering
    \begin{subfigure}{0.25\textwidth}
        \centering
        \includegraphics[width=\linewidth]{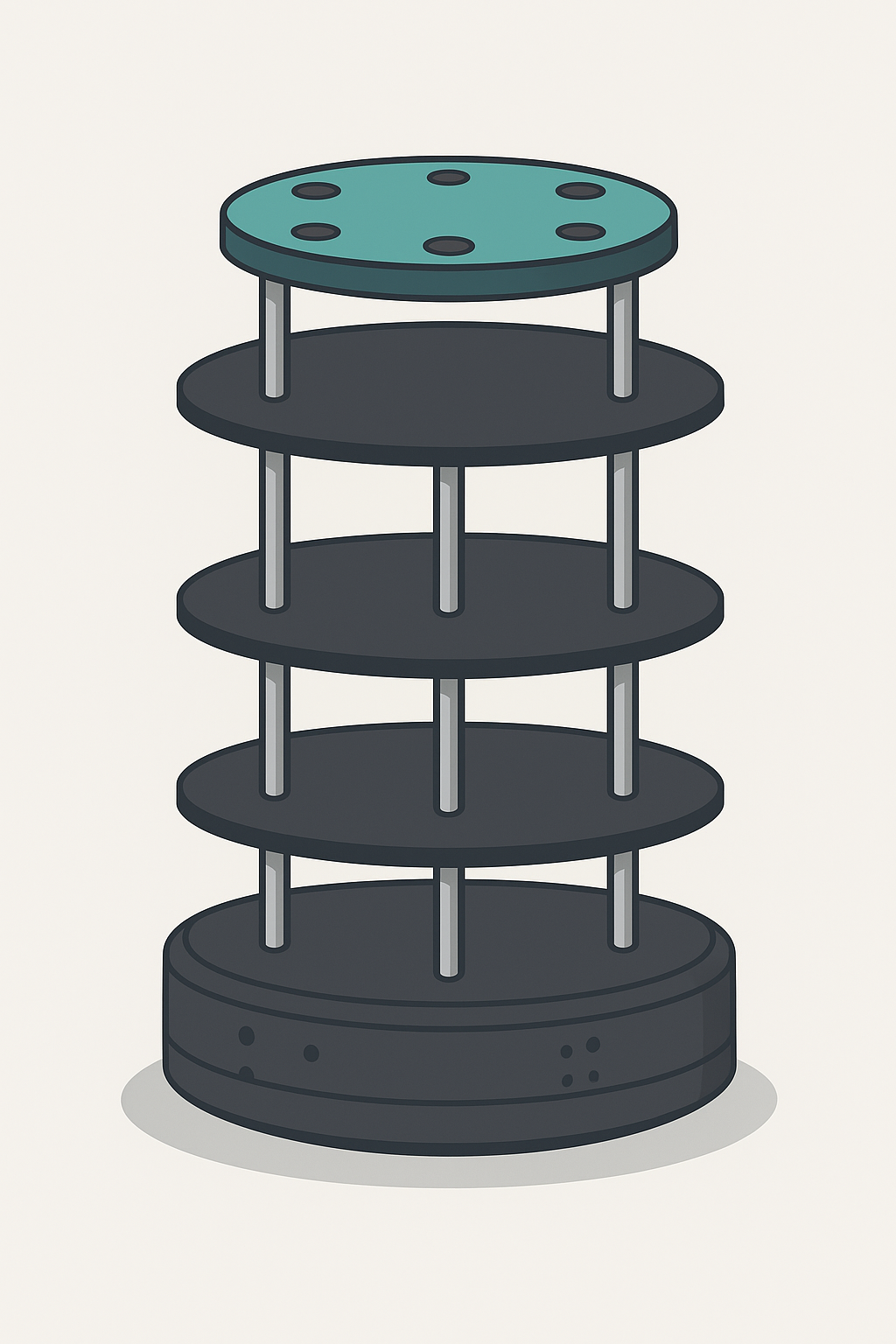}
        \caption{TurtleBot2 with a circular microphone array mounted on top.  }
        \label{fig:turtlebot2}
    \end{subfigure}
    % Second image: Hand Detection by Keypoints
    \begin{subfigure}{0.5\textwidth}
        \centering
        \includegraphics[width=\linewidth]{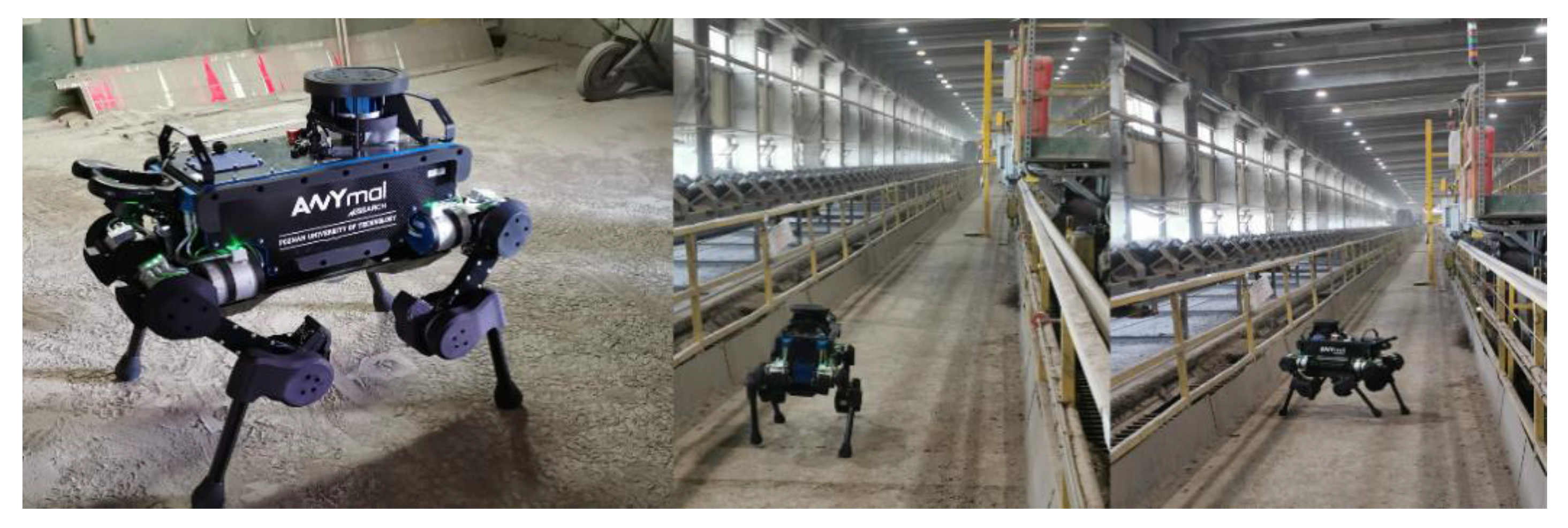}
        \caption{ANYmal quadruped using SSL to inspect a mining belt conveyor \cite{skoczylas2021belt}. Reproduced without modifications from \cite{skoczylas2021belt}, article distributed under the Creative Commons Attribution 4.0 International licence (CC BY 4.0; \url{https://creativecommons.org/licenses/by/4.0/}).}

        \label{fig:anymal}
    \end{subfigure}
    % Third image: Hand Segmentation
    \begin{subfigure}{0.35\textwidth}
        \centering
        \includegraphics[width=\linewidth]{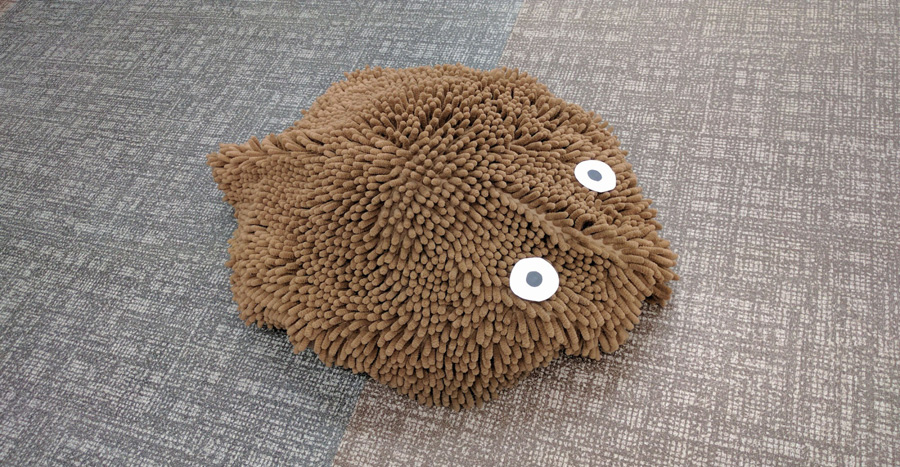}
        \caption{Fur-covered “pet” robot that localizes humans via SSL for social interaction \cite{suzuki2017development}. Reproduced without modifications from \cite{suzuki2017development}, article distributed under the Creative Commons Attribution-NoDerivatives 4.0 International licence (CC BY-ND 4.0; \url{https://creativecommons.org/licenses/by-nd/4.0/}).}
        \label{fig:pet}
    \end{subfigure}
    \begin{subfigure}{0.35\textwidth}
        \centering
        \includegraphics[width=\linewidth]{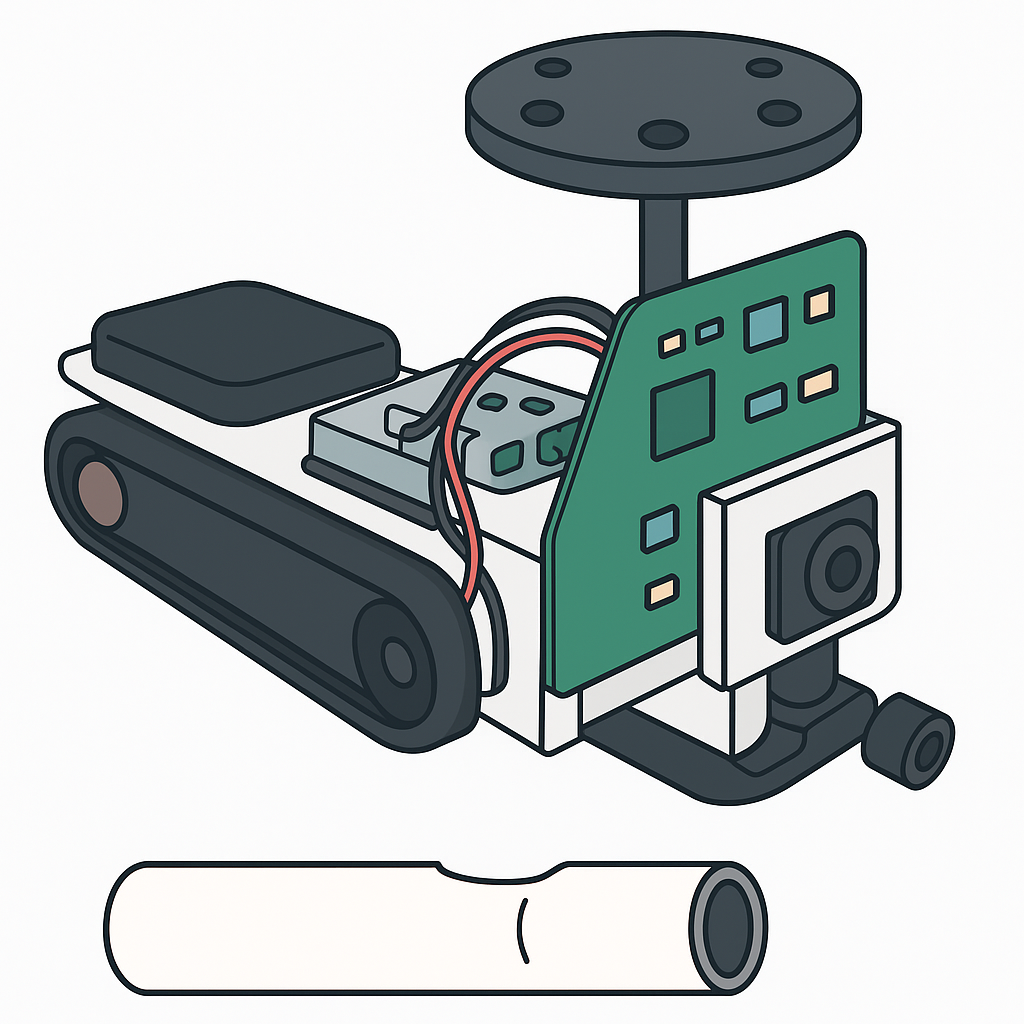}
        \caption{Schematic of the tracked pipe‐inspection robot with a four-microphone array used for sound‐source‐localization‐based defect detection (redrawn and adapted from the concept in \cite{yu2025sparse}).}

        \label{fig:spherical}
    \end{subfigure}
    \begin{subfigure}{0.3\textwidth}
        \centering
        \includegraphics[width=\linewidth]{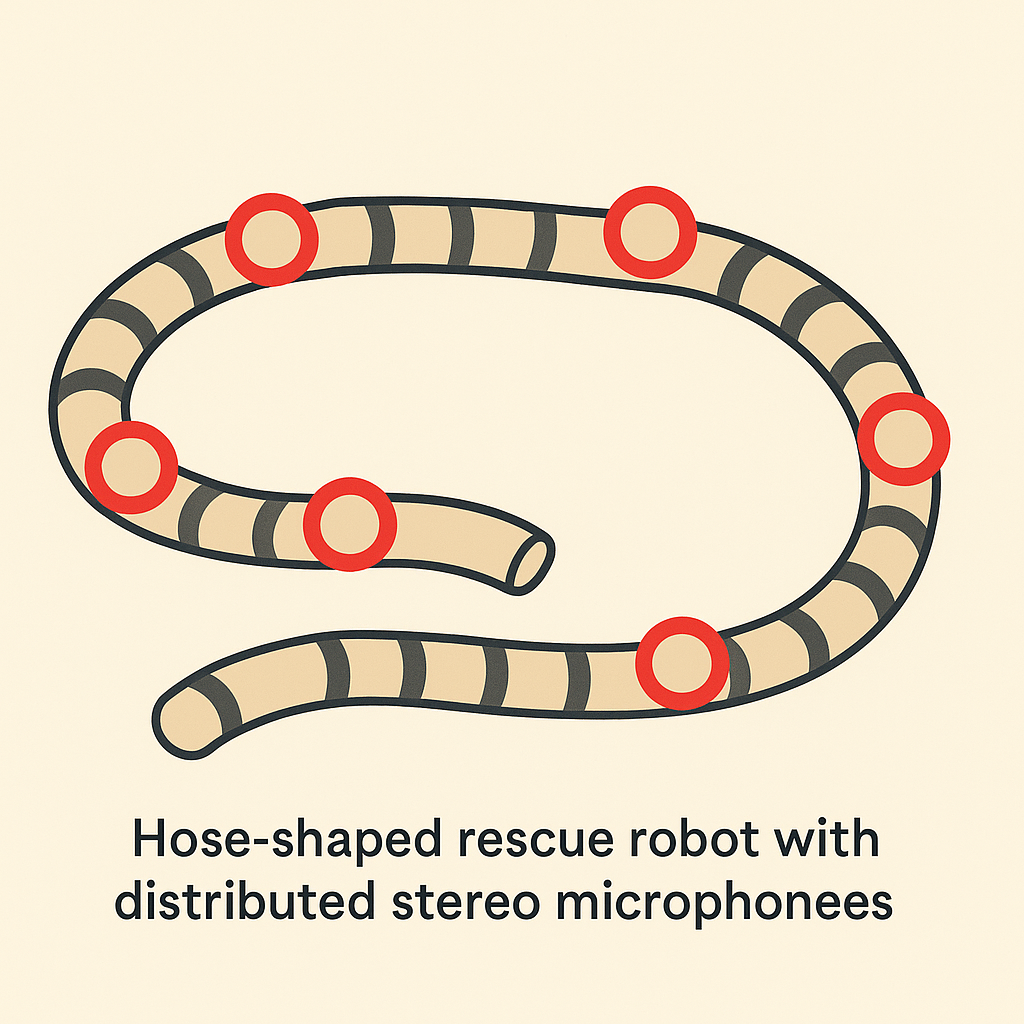}
        \caption{Hose-shaped rescue robot equipped with distributed stereo microphones for sound-source localisation (redrawn and adapted from the concept in \cite{mae2017sound}).}
        \label{fig:hose}
    \end{subfigure}
    \begin{subfigure}{0.3\textwidth}
        \centering
        \includegraphics[width=\linewidth]{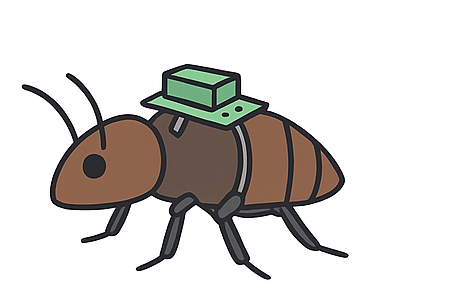}
        \caption{Insect-scale bio-robot equipped with a lightweight microphone array for SSL in search-and-rescue missions, (redrawn and adapted from the concept in \cite{latif2015sound}).}
        \label{fig:biobot_ssl}
    \end{subfigure}
    \caption{Different mobile robot types employed in SSL studies.}
 \label{fig:mobrobot_types}
 
\end{figure*}

\textbf{Mobile Ground Robots:}
Representing the largest category in our review with 27 papers, mobile ground robots are a cornerstone of robotic auditioning systems. These platforms, encompassing wheeled, legged, and tracked designs, leverage SSL to navigate, monitor, and interact with their environment, particularly in scenarios where visual information may be limited or unreliable. The TurtleBot was a commonly utilized mobile ground robot in these studies \cite{nguyen2017long,an2018reflection,boztas2023sound,gala2019realtime,michaud20203d}, while many researchers also employed customized robotic platforms tailored to their specific research objectives, such as ANYmal legged robot \cite{skoczylas2021belt}, a pet robot \cite{suzuki2017development}, pipe inspector robot \cite{yu2025sparse}, a hose-shaped robot \cite{mae2017sound}, and a bio-robot \cite{latif2015sound}. Figure \ref{fig:mobrobot_types} represents these mobile robots.

Microphone arrays are typically mounted on their chassis or integrated into dedicated sensor heads. Their applications include:
\begin{itemize}
    \item \textit{Acoustic Navigation and Mapping:} Mobile robots can utilize SSL to identify and localize fixed sound sources (e.g., specific machinery hums in an industrial setting, ventilation systems, or public address announcements) as acoustic landmarks for Simultaneous Localization and Mapping (SLAM) or precise navigation in GPS-denied or visually ambiguous areas.
    \item \textit{Hazard Detection and Avoidance:} For service robots in public spaces or industrial robots on factory floors, SSL enables the early detection and localization of unexpected or dangerous sounds (e.g., a car horn, breaking glass, a falling object, abnormal machinery sounds). This facilitates proactive collision avoidance or emergency response.
    \item \textit{Sound-Guided Exploration and Search:} In search and rescue scenarios (e.g., navigating collapsed buildings or smoke-filled areas), mobile robots can rely on SSL to precisely locate sounds like human voices or whistles, effectively guiding their exploration towards potential survivors.
    \item \textit{Security and Surveillance:} Mobile robots can patrol large areas, employing SSL to detect and track intruders or suspicious acoustic events (e.g., footsteps, suspicious voices) in various service or security contexts.
\end{itemize}

\begin{figure*}
    
    \centering
    \begin{subfigure}{0.3\textwidth}
        \centering
        \includegraphics[height=3cm]{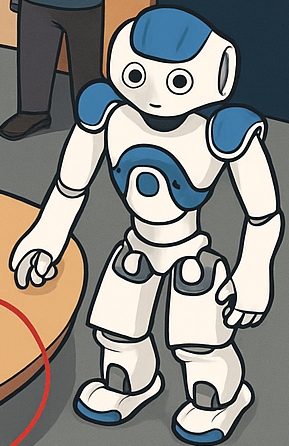}
        \caption{NAO used in \cite{li2016reverberant,takeda2017unsupervised,chen2020efficient,boztas2023sound}.  }
        \label{fig:nao}
    \end{subfigure}
    % Second image: Hand Detection by Keypoints
    \begin{subfigure}{0.3\textwidth}
        \centering
        \includegraphics[height=3cm]{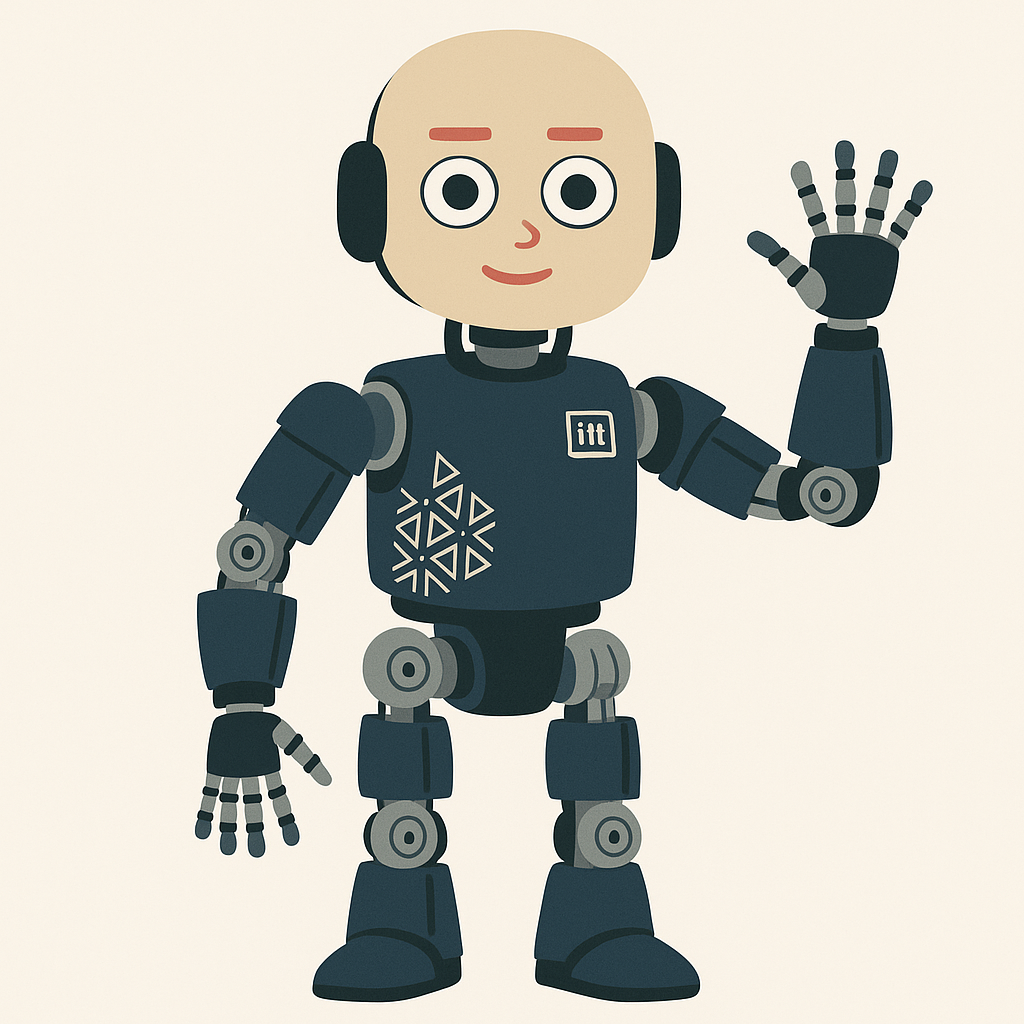}
        \caption{iCub used in  \cite{davila2014improving,davila2018enhanced,kothig2019bayesian,gonzalez2021self}.}

        \label{fig:icub}
    \end{subfigure}
    % Third image: Hand Segmentation
    \begin{subfigure}{0.3\textwidth}
        \centering
        \includegraphics[height=3cm]{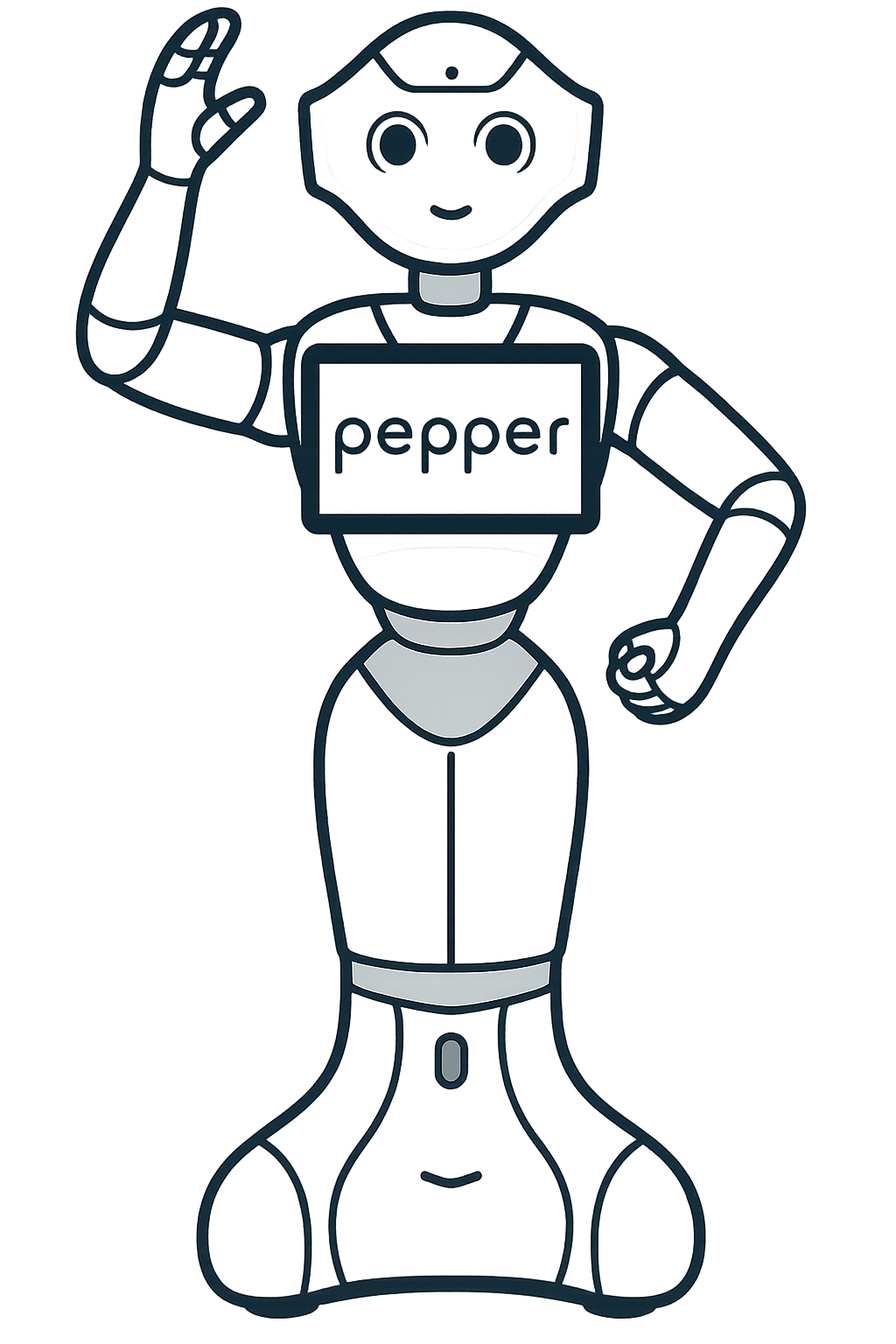}
        \caption{Pepper robot used in \cite{he2018deep,he2021deep}.}
        \label{fig:pepper}
    \end{subfigure}
    \caption{Common humanoids in SSL studies.}
 \label{fig:humanoid_types}
 
\end{figure*}

\textbf{Humanoid Robots:}
Eighteen papers among the 78 reviewed implemented SSL on humanoid robots or platforms specifically designed with a dummy head configuration e.g. Knowles Electronics Mannequin for Acoustic Research (KEMAR), widely used in SSL \cite{keyrouz2014advanced}. Figure \ref{fig:humanoid_types} shows the typical humanoid robots used in SSL. The NAO robot (Figure \ref{fig:nao}) \cite{li2016reverberant,takeda2017unsupervised,chen2020efficient,boztas2023sound} and iCub (Figure \ref{fig:icub}) \cite{davila2014improving,davila2018enhanced,kothig2019bayesian,gonzalez2021self} were frequently employed platforms in these studies, alongside others such as Pepper (Figure \ref{fig:pepper}) \cite{he2018deep,he2021deep}, Hearbo \cite{narang2014auditory} and HRP-2 \cite{asano2015sound}.

These humanoid robots, designed to interact with and operate in human-centric environments, benefit immensely from sophisticated auditory perception. Equipped with microphone arrays, often integrated into their heads (mimicking binaural hearing) or across their torsos, they utilize SSL for:
\begin{itemize}
    \item \textit{Natural Human-Robot Interaction (HRI):} Localizing the direction of human speech allows humanoids to naturally orient their head and gaze towards a speaker, facilitating engaging conversations and conveying attentiveness. This is crucial for social and companion robots aiming to build rapport with users.
    \item \textit{Multi-Speaker Tracking:} In dynamic social settings, humanoids can employ advanced SSL techniques (e.g., using deep learning or subspace methods) to identify and track multiple simultaneous speakers, enabling complex multi-party conversations and selective listening.
    \item \textit{Enhanced Situational Awareness:} Beyond direct interaction, SSL enables humanoids to detect and localize various environmental sounds, such as alarms, doorbells, or footsteps, contributing to their overall understanding of the surrounding service environment.
\end{itemize}

\textbf{Unmanned Aerial Vehicles (UAVs):}
eleven papers in our review specifically utilized UAVs for their SSL investigations. UAVs present unique challenges and opportunities for SSL due to their inherent ego-noise (from propellers) and often limited payload capacity. Despite these hurdles, SSL offers them extended perception capabilities:
\begin{itemize}
    \item \textit{Remote Monitoring and Surveillance:} UAVs equipped with SSL can perform long-range acoustic monitoring for environmental applications (e.g., detecting illegal logging, tracking wildlife based on vocalizations) or security tasks (e.g., localizing gunshots or human activity over large, inaccessible terrains).
    \item \textit{Search and Rescue in Challenging Terrains:} In search and rescue operations over vast or difficult-to-traverse areas (e.g., dense forests, mountainous regions), UAVs can use SSL to pinpoint distress calls or specific human sounds, directing ground teams more efficiently.
    \item \textit{Hazard Identification:} Detecting and localizing critical sounds like explosions, gas leaks (through associated sounds), or structural failures from a safe distance in industrial or disaster zones.
    \item \textit{Traffic Monitoring:} In service or urban planning contexts, UAVs can localize vehicles based on their sounds, contributing to traffic flow analysis.
\end{itemize}

\subsection{Categorization by Application Domain}

\begin{figure}[htbp]
    \centering
    \includegraphics[width=0.5\textwidth]{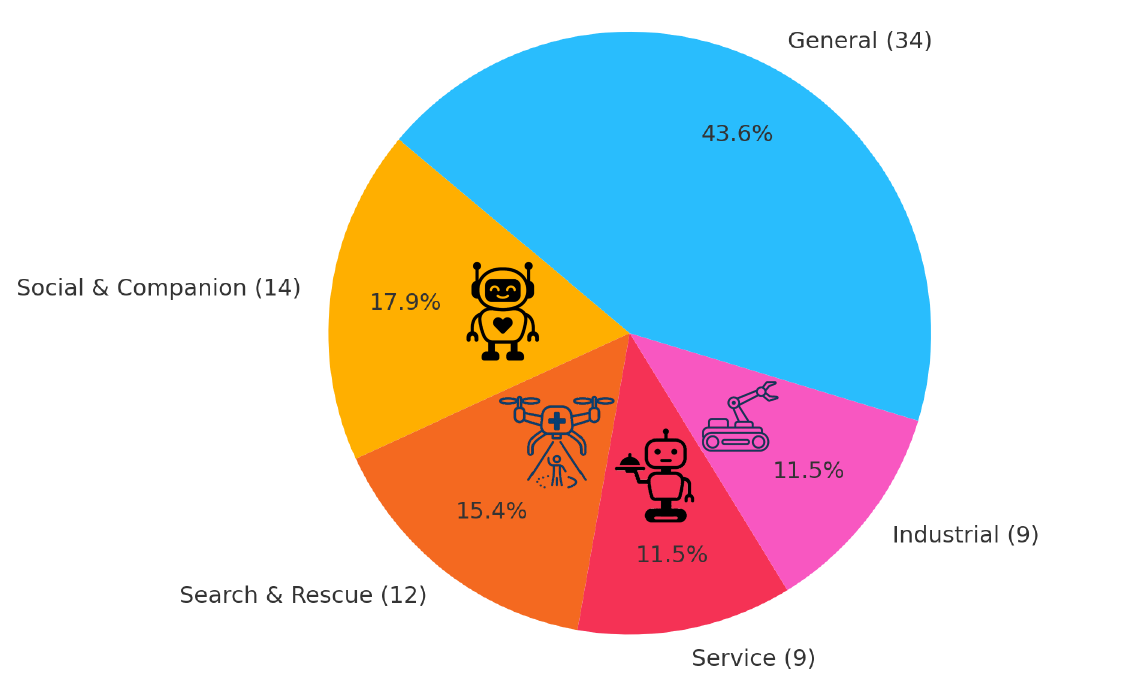} % Or .pdf, .jpg etc.
    \caption{Distribution of SSL applications in robotics across different domains in our review.}
    \label{fig:domain_pie_chart} % This is the label you'd use in \ref{}
\end{figure}

While the robot type defines the platform, the application domain dictates the specific tasks and environmental challenges SSL must address. Based on our review, four primary domains have emerged: social and companion, search and rescue, service, and industrial robotics. We also introduce a fifth category, "General" for studies that do not restrict their applications to a single specific domain. The proportion of reviewed papers falling into each domain is depicted in a pie chart in Figure \ref{fig:domain_pie_chart}.

\textbf{Social and Companion:}
Among the reviewed literature, fourteen studies focused on enhancing social interaction and companionship through robotic systems. These works primarily employed humanoid robots \cite{davila2014improving,kim2015improved,li2016reverberant,takeda2017unsupervised,he2018deep,pang2019multitask,davila2018enhanced,chen2020efficient,shi2023audio}, although some mobile ground robots were also utilized \cite{korayem2021design,shi2023audio}. For robots designed to interact with and assist humans in social contexts, SSL is paramount for fostering natural and engaging relationships. It enables accurate speech localization for fluent conversations, allows the robot to discern who is speaking in group settings, and facilitates attentive behaviors by orienting towards auditory cues, thereby enhancing the robot's perceived intelligence and empathy (through possible facial analysis as an additional modality \cite{korayem2021design,shi2023audio}). A central theme in these studies is the detection and localization of human speech, often considering variations such as male and female speakers due to the human-centric nature of the sound sources \cite{takeda2017unsupervised,pang2019multitask}. Notably, an interesting research by Chen et al. \cite{chen2020efficient} explored localizing the referee whistle in RoboCup competitions for NAO robots, requiring distinction from other ambient sounds like human speech and claps.

\textbf{Search and Rescue:}
Twelve papers in our review addressed the theme of search and rescue, a domain where SSL serves as a vital, often life-saving, capability for robots deployed in hazardous or disaster-stricken areas where visibility is severely compromised. Sound separation and detection are critical factors in this domain, as the robot must reliably distinguish target sounds from its own ego-noise and other non-pertinent ambient noises. These robots leverage SSL to pinpoint the exact locations of trapped survivors by detecting specific sounds, such as human speech and chirps \cite{manamperi2022drone,strauss2018dregon} or whistles \cite{hoshiba2017design,nakadai2017development}, thereby guiding rescue teams to precise areas within debris fields or collapsed structures. Most of these studies UAVs in both outdoor \cite{hoshiba2017design,nakadai2017development,yamamoto2024implementation,clayton2023embedded,yamada2023placement} and indoor settings \cite{strauss2018dregon,manamperi2022drone}. Beyond UAVs, some research also incorporated mobile ground robots \cite{zhang2019sound} and humanoids \cite{asano2015sound}, as well as custom-designed robots like hose-shaped or bio-inspired platforms for specialized search and rescue experiments \cite{mae2017sound,latif2015sound}.

\textbf{Service:}
Nine studies among the reviewed papers focused on robotic applications within the service domain. SSL empowers service robots to operate effectively in diverse human environments, ranging from smart homes to public spaces. Key applications include speaker localization for intuitive interaction, the detection of specific household sounds (e.g., door knocks, appliance alarms, and hairdryers \cite{grondin2016noise,kim2021mimo}) or public place acoustics (e.g., in classrooms \cite{yook2015fast}), and the monitoring for unusual acoustic events in residential or commercial settings, including security applications \cite{han2020research}. These studies primarily focused on indoor environments, predominantly utilizing ground mobile robots \cite{grondin2016noise,kim2021mimo,chen2022large,chen2022broadband,gamboa2022real,jo2025sound}.

\textbf{Industrial:}
Among the reviewed articles, nine papers specifically addressed the industrial domain. Six of these studies directly focused on industrial condition monitoring for fault detection using SSL with robotic platforms \cite{song2020automatic,shiri2021inspection,skoczylas2021belt,lv2024motor,lv2024overview,yu2025sparse}. Complementing these, three other papers, while not directly implementing robots in their experimental setups, explored SSL within contexts highly relevant to future robotic integration in industrial environments. Jalayer et al. \cite{jalayer2023conceptual,jalayer2024convlstm} investigated applications in manufacturing settings, considering both machinery sounds and human speech, thereby laying groundwork for robots operating alongside human workers. With the same spirit, Sun et al. \cite{sun2017indoor} discussed the potential for industrial SSL implementation by utilizing common sounds in such environments, including human speech, machinery sounds, and telephone rings in indoor settings. Although SSL remains relatively underexplored in this domain compared to others, manufacturing plants, warehouses, and other industrial settings hold immense potential for robots equipped with advanced SSL techniques to significantly enhance automation and safety. Robots can leverage SSL for predictive maintenance by localizing anomalous machinery sounds (e.g., grinding, squealing) to identify failing components, monitor for critical safety alarms, or precisely track the movement of industrial vehicles and personnel for improved collision avoidance and operational efficiency.

\section{Challenges and Future Avenues}
\label{sec:Challenges and Future Avenues}

Despite significant advances in SSL for robotics, particularly with the integration of deep learning methods, numerous challenges remain. This section examines the current limitations and emerging research directions that aim to address these challenges and further advance the field.

\subsection{Current Challenges}

\subsubsection{Environmental Robustness}
Although compared to traditional signal processing, DL models performed better in complex environments, achieving robust and reliable performance across diverse and highly dynamic acoustic environments is still a hurdle for robot audition systems. These challenges can be categorized as follows:

\begin{itemize}
\item \textbf{Extreme Reverberation:} In large indoor spaces with reflective surfaces, extreme reverberation severely degrades a robot's ability to precisely map its acoustic surroundings. This phenomenon continues to be a significant obstacle to SSL accuracy in multi-source and noisy environments, causing sound reflections to confuse the robot's directional cues.

\item \textbf{Different noise types:} Real-world scenarios are characterized by complex, unpredictable background noise, such as human chatter, machinery, or general environmental sounds, which can mask critical acoustic cues. Furthermore, a robot's own self-generated noise (ego-noise) from motors, actuators, and movement significantly complicates its internal auditory focus. Although many studies considered ego-noise in their studies and datasets \cite{strauss2018dregon}, considering multiple noise types and their cumulative effect that can happen in realistic applications is still a challenge. Also, considering the effect of microphone internal noise (caused by imperfection \cite{jalayer2024convlstm}) could add complexity, and this should be addressed since microphones do not always work perfectly.

% Rascon and Meza \cite{rascon2020survey} highlight that robustness against such noise remains a critical challenge for SSL in robotics.
\item \textbf{Dynamic Acoustic Conditions and Unstructured Environments:} As robots navigate, acoustic conditions continuously change due to varying listener positions relative to sources and reflections. Adapting to these dynamic changes without requiring re-calibration or extensive retraining presents a significant challenge for both traditional and deep learning methods. Moreover, extending SSL approaches to outdoor (especially for UAVs) or highly unstructured environments introduces additional complexities related to wind, varying atmospheric conditions, and unpredictable sound propagation.
\end{itemize}

\subsubsection{Multi-Source Scenarios}
Accurately localizing and distinguishing multiple simultaneous sound sources, particularly in cluttered acoustic environments, poses a significant scene analysis problem for robotic systems:
\begin{itemize}
\item \textbf{Source Separation vs. Localization Interdependence:} A fundamental dilemma for a robot's auditory processing pipeline is the interdependency between source separation and localization. Effective separation of individual sound streams often requires prior knowledge of source locations, while accurate localization may require separated or enhanced source signals, creating a "chicken-and-egg" problem for the robot's acoustic intelligence.
In this regard, some previous SSL studies implemented separate DL models for sound counting, classification and localization tasks \cite{kim2011direction}, while some did all within one model \cite{he2018deep}.
\item \textbf{Overlapping Sources:} 
 When multiple sound sources extensively overlap in time and frequency, it profoundly challenges a robot's ability to differentiate and pinpoint distinct acoustic events. This is particularly difficult for speech sources with similar spectral characteristics or when multiple human speakers are closely spaced (a cocktail party scenario \cite{shi2023audio}), affecting a robot's ability to focus on a specific speaker.
\item \textbf{Variable Source Numbers:}  Real-world scenarios involve a fluctuating number of active sound sources over time. Developing robotic auditory systems that can dynamically adapt to changing numbers of sources without explicit prior knowledge remains a complex task, requiring flexible and scalable processing architectures.
\item \textbf{Source Tracking and Identity Maintenance:} 
Maintaining consistent identity tracking of multiple moving sound sources over extended periods \cite{wang2023multiple}, especially through instances of silence, occlusion, or signal degradation, presents significant difficulties that current robotic audition methods have not fully resolved \cite{grumiaux2022survey}. This capability is essential for a robot to maintain situational awareness and interact intelligently with dynamic entities.
\end{itemize}

\subsubsection{Practical Implementation Constraints}
Deploying robust SSL systems on real robotic platforms introduces several practical constraints that must be overcome:
\begin{itemize}
\item \textbf{Computational Efficiency and Power Consumption:} 
Real-time processing requirements coupled with the limited computational resources and strict power budgets on many robotic platforms directly impact a robot's operational endurance and real-time responsiveness. There is a growing interest in "energy-efficient wake-up technologies" \cite{yang2023acoustic} for SSL, as highlighted by Khan et al. \cite{khan2025review}, to enable long-duration missions for battery-powered robots by minimizing energy expenditure on continuous acoustic monitoring.
\item \textbf{Microphone Array Limitations and Flexibility:} Physical constraints on microphone placement, quality, and array geometry across diverse robotic platforms (e.g., humanoid heads, mobile robot chassis, UAVs) impose inherent sensory limitations on a robot's acoustic perception and can significantly impact SSL performance.
The type of input features and microphone array configurations heavily influence the effectiveness of deep learning approaches, often requiring model retraining for different robot setups. Moreover, supporting dynamic microphone array configurations (e.g., on articulated or reconfigurable robots) remains a challenge.
\item \textbf{Calibration and Maintenance:} Ensuring consistent acoustic performance over a robot's operational lifespan demands robust self-calibration routines within its perceptual architecture to address issues such as microphone drift, physical damage, or changes in robot configuration that may affect the acoustic properties of the system. This is crucial for maintaining the integrity of the robot's auditory data.
\end{itemize}

\subsubsection{Data and Learning Challenges}
Deep learning approaches, while powerful, face specific challenges related to data management and learning processes for SSL in robotics:
\begin{itemize}
\item \textbf{Limited Training Data and Annotation Complexity:} 
% Collecting diverse, high-quality labeled acoustic data for robot SSL is inherently time-consuming and expensive. This is particularly true for robotic applications, where datasets should ideally cover various robot configurations, acoustic environments, and complex real-world interactions, restricting a robot's ability to generalize its learned auditory models \cite{rascon2017localization}.
Collecting diverse, high-quality labeled acoustic data for SSL is inherently time-consuming and expensive for robotic applications. Unlike vision datasets (where images can often be manually labeled), obtaining the true direction or position of a sound source at each time requires specialized equipment (e.g., motion tracking systems as used in LOCATA \cite{evers2020locata}) or careful calibration with known source positions. This becomes even harder if either the sound source or the robot (or both) are moving. As a result, truly comprehensive spatial audio datasets for robotics are scarce. Furthermore, a significant impediment is that most research studies often do not share their collected data, hindering broader research progress and comparative analysis.
\item \textbf{Lack of Comprehensive Benchmarks in SSL:} Despite some efforts (e.g., CASE challenges in the last decade \cite{mesaros2025decade}), there is a notable lack of comprehensive, widely adopted benchmark datasets (like ImageNet in visual object recognition) in the field of SSL. This absence makes it difficult for researchers to uniformly test and compare the performance of different models, impeding standardized evaluation and progress towards common goals.
\item \textbf{Sim-to-Real Gap and Generalization:} While simulated environments, such as Pyroomacoustics \cite{scheibler2018pyroomacoustics} and ROOMSIM \cite{campbell2005matlab,schimmel2009fast}, can generate large amounts of training data, bridging the fidelity gap between simulated and real-world acoustics remains a critical barrier to a robot's transition from simulated training to autonomous real-world operation. Consequently, models trained on specific environments or conditions often fail to generalize robustly to new, unseen scenarios, limiting their real-world applicability for robotic deployment.
\item \textbf{Interpretability:} The "black box" nature of many deep learning models complicates debugging, understanding their decision-making processes, and ensuring their reliability. This lack of interpretability can be particularly problematic for safety-critical robotics (e.g., rescue robots \cite{nakadai2017development} or condition monitoring robots \cite{lv2024motor}) applications where explainability and a human operator's trust in the robot's auditory decisions are paramount.
\item \textbf{Accurate 3D Localization and Distance Estimation:} While 2D Direction of Arrival (DoA) is commonly addressed, accurately estimating the distance to a sound source, especially in reverberant environments, remains a more complex task for a robot's spatial awareness. As highlighted by Rascon et al. \cite{rascon2017localization}, reliable 3D localization (azimuth, elevation, and distance) with high resolution is essential for a robot's precise navigation, manipulation, and interaction within a volumetric space, but it is not yet consistently achievable in real-time.
\end{itemize}

\subsection{Future Opportunities and Avenues}

Building upon the current challenges, the field of robotic sound source localization presents numerous exciting opportunities for research and development. These avenues aim to elevate a robot's auditory intelligence, enabling more autonomous, perceptive, and adaptable machines for a wide range of applications.

\subsubsection{Enhancing Robustness and Adaptability in Robot Audition}
Future work will focus on equipping robots with auditory systems capable of navigating the most challenging acoustic environments with unprecedented reliability:
\begin{itemize}
\item \textbf{Adaptive Noise and Reverberation Suppression:} Developing advanced signal processing and deep learning techniques that can adaptively suppress diverse non-stationary noise and mitigate extreme reverberation in real-time. This includes research into robust ego-noise cancellation specific to robotic platforms, ensuring a robot can maintain auditory focus even during rapid movement or noisy operations.
\item \textbf{Outdoor and Unstructured Acoustic Modeling:} Expanding SSL research beyond controlled indoor environments of the labs to truly unstructured outdoor settings. This involves developing new models for sound propagation in open air, by introducing novel outdoor acoustic simulators, accounting for environmental factors like wind, and leveraging multi-modal fusion with visual or inertial sensors to enhance robustness where audio cues might be ambiguous.
\item \textbf{Dynamic Acoustic Scene Understanding:} Moving beyond static snapshot localization to continuous, real-time comprehension of evolving acoustic scenes. This includes rapid adaptation to changing sound source characteristics, varying background noise, and dynamic room acoustics as the robot moves, fostering a more fluid and context-aware auditory perception.
\end{itemize}

\subsubsection{Advanced Multi-Source Auditory Scene Analysis}
Opportunities abound in enabling robots to dissect complex auditory environments with multiple, interacting sound sources:
\begin{itemize}
\item \textbf{Integrated Sound Event Localization and Detection (SELD) with Source Separation:} Developing holistic systems that simultaneously detect, localize, and separate multiple overlapping sound events. This integrated approach, often framed as a multi-task learning problem, promises to break the current "chicken-and-egg" dilemma between separation and localization, allowing robots to extract distinct auditory streams for specific analysis or interaction.
\item \textbf{Robust Multi-Object Acoustic Tracking:} Advancing algorithms for reliable and persistent tracking of multiple moving sound sources, including handling occlusions, disappearances, and re-appearances. This is crucial for collaborative robots interacting with multiple humans or for inspection robots monitoring various machinery components in parallel.
\item \textbf{Dynamic Source Counting and Characterization:} Equipping robots with the ability to dynamically estimate the number of active sound sources in real-time, along with their types (e.g., speech, machinery, alarms). This capability enhances a robot's contextual awareness, allowing it to prioritize relevant acoustic information within a complex environment.
\end{itemize}

\subsubsection{Efficient and Flexible Robotic System Integration}
Future research will drive the development of SSL solutions optimized for practical deployment on diverse robotic hardware:
\begin{itemize}
\item \textbf{Lightweight and Energy-Efficient Deep Learning Models:} Designing novel, compact deep learning architectures and employing techniques like model quantization, pruning, and knowledge distillation tailored for execution on resource-constrained edge AI platforms. Also, advancement in power saving techniques (e.g. wake-up strategy described in Khan et al. survey \cite{khan2025review}) promise to keep the auditory system off until an acoustic event of interest occurs, allowing the robot to reserve precious energy for localization and task execution.
This will enable robots to perform complex SSL tasks while maintaining long operational durations and reducing power consumption.
\item \textbf{Flexible and Self-Calibrating Microphone Arrays:} Investigating SSL methods that are robust to variations in microphone array geometry, supporting dynamic reconfigurations inherent to mobile or articulated robots. Furthermore, developing autonomous, self-calibrating routines for microphone arrays will reduce deployment complexity and ensure consistent performance over a robot's lifespan, compensating for sensor drift or minor physical changes. Techniques like Non-Synchronous Measurement (NSM) technology, which allows a small moving microphone array to emulate a larger static one, offer promising avenues for achieving high-resolution localization with fewer microphones while managing the challenges of dynamic array configurations \cite{lv2024motor}.
\item \textbf{Hardware-Software Co-Design for Robot Audition:} Fostering a holistic approach where microphone array design, sensor placement on the robot, and signal processing algorithms are jointly optimized. This co-design can lead to highly efficient and purpose-built auditory systems that maximize performance within a robot's physical and computational constraints.
\end{itemize}

\subsubsection{Advancements in Data-Driven Learning and Robot Intelligence}
Significant opportunities lie in overcoming data limitations and enhancing the intelligence and transparency of a robot's acoustic learning processes:
\begin{itemize}
\item \textbf{Large-Scale, Diverse, and Shared Datasets:} A critical opportunity is the creation and broad dissemination of large-scale, diverse, and meticulously annotated benchmark datasets specifically for robotic SSL. These datasets should include challenging real-world scenarios, diverse robot platforms, varying microphone array configurations (including dynamic ones), and precise ground truth for moving sources and robots. Initiatives to encourage data sharing across research institutions are vital to foster collaborative progress and provide standardized benchmarks for model comparison.
\item \textbf{Unsupervised, Self-Supervised, and Reinforcement Learning for SSL:} Exploring learning paradigms that minimize the reliance on labor-intensive labeled data. This includes techniques that allow robots to learn spatial acoustic features from vast amounts of unlabeled audio data gathered during operation, as well as reinforcement learning approaches where a robot can optimize its SSL performance through interaction with its environment \cite{grumiaux2022survey,jekaterynczuk2023survey}.
\item \textbf{Bridging the Sim-to-Real Gap with Domain Adaptation:} Developing robust domain adaptation and transfer learning techniques to effectively transfer knowledge from simulated acoustic environments to real-world robotic deployment. This will involve more sophisticated acoustic simulations that accurately model complex reverberation and noise, coupled with techniques that make deep learning models more resilient to the inevitable discrepancies between simulated and real data \cite{zhang2025sound}.
% \item \textbf{Physics-Informed Deep Learning and Hybrid Models:} Integrating fundamental principles of acoustics and signal processing into deep learning architectures. This "physics-informed" approach can lead to more data-efficient, generalizable, and interpretable models, combining the strengths of model-based methods with the learning power of neural networks \cite{grumiaux2022deep, khan2024sound}.
\item \textbf{Foundation Models for Semantic Interpretation:} A promising avenue involves integrating Large Language Models (LLMs) into the robot's auditory processing pipeline, particularly after successful sound source localization and speech recognition. Once an SSL system pinpoints the origin of human speech, and an Automatic Speech Recognition (ASR) system transcribes it, LLMs can be employed to provide deeper contextual understanding and semantic interpretation of verbal commands, queries, or intentions. This would enable robots to engage in more natural, nuanced, and multi-turn dialogues, disambiguate vague instructions based on conversational history or common sense, and ground abstract concepts in the robot's physical environment, moving beyond mere utterance processing to true linguistic comprehension.
\item \textbf{Hybrid Models and Multi-Modal Fusion for Holistic Perception:} Future research will increasingly focus on hybrid models that intelligently combine deep learning with traditional signal processing techniques, and crucially, on fusing SSL outputs with other sensory modalities. This includes tightly coupled integration of acoustic data with visual information from cameras, e.g., for audio-visual speaker tracking, identifying sounding objects, human facial expression in HRI \cite{korayem2021design,shi2023audio}, and haptic feedback (e.g., for contact localization on robot limbs). Such multi-modal fusion allows robots to build a more comprehensive and robust perception of their environment, compensating for limitations in any single modality and enabling richer semantic understanding of the auditory scene.
\item \textbf{Explainable AI (XAI) for Robot Audition:} Research into methods that allow deep learning-based SSL systems to provide transparent explanations for their localization decisions. This will enhance human operators' trust in autonomous robots and facilitate debugging and refinement of auditory perception systems in safety-critical applications.
% \item \textbf{Accurate 3D Localization and Distance Estimation:} Further advancing techniques for robust 3D SSL, with a particular focus on consistently and accurately estimating the distance to a sound source in real-time, even in reverberant environments. This capability is paramount for robots performing precise manipulation, navigation, and human-robot collaboration within a volumetric space \cite{rascon2020survey}.
\end{itemize}

\section{Conclusions}
\label{sec:conclusions}

Sound source localization (SSL) is a pivotal capability for enhancing robot autonomy, facilitating seamless human-robot interaction, and enabling intelligent environmental awareness. This review has provided a comprehensive synthesis of SSL in robotics, with a particular focus on the transformative impact of deep learning methods over the past decade.
% We began by revisiting the fundamental principles of SSL, highlighting its unique advantages over other localization modalities, such as its robustness to occlusions and darkness. We then transitioned into a detailed exploration of deep learning architectures, from foundational CNNs and RNNs to sophisticated attention-based models, and examined how these data-driven approaches implicitly handle complex acoustic phenomena like reverberation and noise—challenges that traditionally limited classical methods. The critical role of data and training strategies, including synthetic data generation and various learning paradigms, was also thoroughly discussed.
We began by revisiting the fundamental principles of SSL and the traditional methods that formed its bedrock, such as Time Difference of Arrival (TDOA), beamforming, Steered-Response Power (SRP), and subspace analysis methods, highlighting their core mechanisms and inherent limitations. We then transitioned into a comprehensive exploration of deep learning architectures, from foundational Machine Learning, shallow neural networks, Convolutional Neural Networks (CNNs), and Recurrent Neural Networks (RNNs) to sophisticated attention-based models. We examined how these data-driven approaches implicitly handle complex acoustic phenomena like reverberation and noise—challenges that traditionally limited classical methods. The critical role of data and training strategies, including synthetic data generation and various learning paradigms, was also thoroughly discussed.
Furthermore, we mapped the diverse applications of SSL in robotics across various robotic types (i.e., mobile ground robots, humanoids, UAVs) and domains, including social and companion, search and rescue, service, and industrial. This provided a practical overview of how SSL is being integrated into real-world robotic systems, ranging from speech command recognition to anomaly detection and situational awareness.

Despite significant advancements, particularly with deep learning, several key challenges remain. Achieving environmental robustness in highly reverberant, noisy, and dynamic conditions, especially in unstructured outdoor environments, continues to demand innovative solutions. The accurate localization and tracking of multiple, overlapping sound sources remain complex, requiring better source separation and identity maintenance capabilities. Practical implementation constraints such as computational efficiency, power consumption, and the inherent limitations of microphone arrays on robotic platforms necessitate the development of lightweight, energy-efficient algorithms and flexible hardware designs. Lastly, data and learning challenges, including the scarcity of diverse, labeled datasets, the "sim-to-real" gap, and the current lack of comprehensive benchmarks, hinder the generalization and real-world deployability of deep learning models. The black-box nature of many deep learning approaches also poses interpretability challenges, particularly for safety-critical robotic applications.

Looking forward, the future of robotic SSL is rich with promising opportunities. Developing adaptive and robust auditory systems that can autonomously cope with extreme acoustic conditions will be paramount. Advancements in multi-source auditory scene analysis, combining localization with intelligent source separation and tracking, will unlock new levels of robot perception in complex social and industrial settings. Crucially, the integration of foundation models, such as Large Language Models, with SSL outputs presents a transformative avenue for robots to not only pinpoint sound sources but also semantically interpret human speech, enabling more natural and intelligent human-robot interaction. Similarly, hybrid models and multi-modal fusion with visual and haptic promise to yield a more holistic and robust understanding of the environment. Overcoming data limitations through unsupervised and self-supervised learning, coupled with the creation of large-scale, shared benchmark datasets for robotic contexts, will accelerate progress. Finally, innovations in efficient hardware-software co-design and explainable AI will pave the way for deployable, trustworthy, and energy-efficient SSL solutions for the next generation of autonomous robots.
By continuing to bridge the gap between acoustic signal processing, advanced deep learning, and robotic system integration, the field is poised to equip robots with auditory capabilities that rival, and in some aspects even surpass, human hearing in specific operational contexts. This will enable robots to become more perceptive, interactive, and intelligent agents in our increasingly complex world.

\bibliography{biblio}

\end{document}